\pdfoutput=1

\documentclass[11pt]{article}

\usepackage[final]{acl}

\usepackage{times}
\usepackage{latexsym}

\usepackage[T1]{fontenc}

\usepackage[utf8]{inputenc}

\usepackage{microtype}

\usepackage{inconsolata}

\usepackage{booktabs}
\usepackage{colortbl}
\usepackage{multirow}
\usepackage{graphicx}
\usepackage[normalem]{ulem}
\useunder{\uline}{\ul}{}
\usepackage{subcaption}

\usepackage{amsmath, amssymb}
\usepackage[ruled,linesnumbered]{algorithm2e}

\title{Improving Low-Resource Sequence Labeling with Knowledge Fusion and Contextual Label Explanations}

\author{
 \textbf{Peichao Lai\textsuperscript{1}},
 \textbf{Jiaxin Gan\textsuperscript{3}},
 \textbf{Feiyang Ye\textsuperscript{3}},
 \textbf{Wentao Zhang\textsuperscript{1}}, \\
 \textbf{Fangcheng Fu\textsuperscript{1}},
 \textbf{Yilei Wang\textsuperscript{3, \dag}},
 \textbf{Bin Cui\textsuperscript{1,2,*}}
 \\
 \textsuperscript{1}School of Computer Science, Peking University, \\
 \textsuperscript{2}Institute of Computational Social Science, Peking University (Qingdao), \\
 \textsuperscript{3}College of Computer and Data Science, Fuzhou University \\
 \small{
  \textsuperscript{*}\href{mailto: bin.cui@pku.edu.cn}{bin.cui@pku.edu.cn}
  \textsuperscript{\dag}\href{mailto: yilei@fzu.edu.cn}{yilei@fzu.edu.cn}
 }
}

\begin{document}
\maketitle
\begin{abstract}
  Sequence labeling remains a significant challenge in low-resource, domain-specific scenarios, particularly for character-dense languages. Existing methods primarily focus on enhancing model comprehension and improving data diversity to boost performance. However, these approaches still struggle with inadequate model applicability and semantic distribution biases in domain-specific contexts. To overcome these limitations, we propose a novel framework that combines an LLM-based knowledge enhancement workflow with a span-based Knowledge Fusion for Rich and Efficient Extraction (KnowFREE) model\footnote{Our code: https://github.com/aleversn/KnowFREE}. Our workflow employs explanation prompts to generate precise contextual interpretations of target entities, effectively mitigating semantic biases and enriching the model's contextual understanding. The KnowFREE model further integrates extension label features, enabling efficient nested entity extraction without relying on external knowledge during inference. Experiments on multiple domain-specific sequence labeling datasets demonstrate that our approach achieves state-of-the-art performance, effectively addressing the challenges posed by low-resource settings.
\end{abstract}

\section{Introduction}

Sequence labeling is a fine-grained information extraction (IE) task that includes sub-tasks such as named entity recognition (NER), word segmentation, and part-of-speech (POS) tagging, playing a critical role in various downstream natural language processing (NLP) applications.


In low-resource scenarios, sequence labeling remains a persistent challenge, primarily due to the scarcity of domain-specific data, which limits the model's capacity to learn accurate label distributions. Moreover, character-dense languages such as Chinese pose additional difficulties, as the absence of explicit word boundaries greatly complicates label inference.

\begin{figure}[t!]
  \includegraphics[width=\columnwidth]{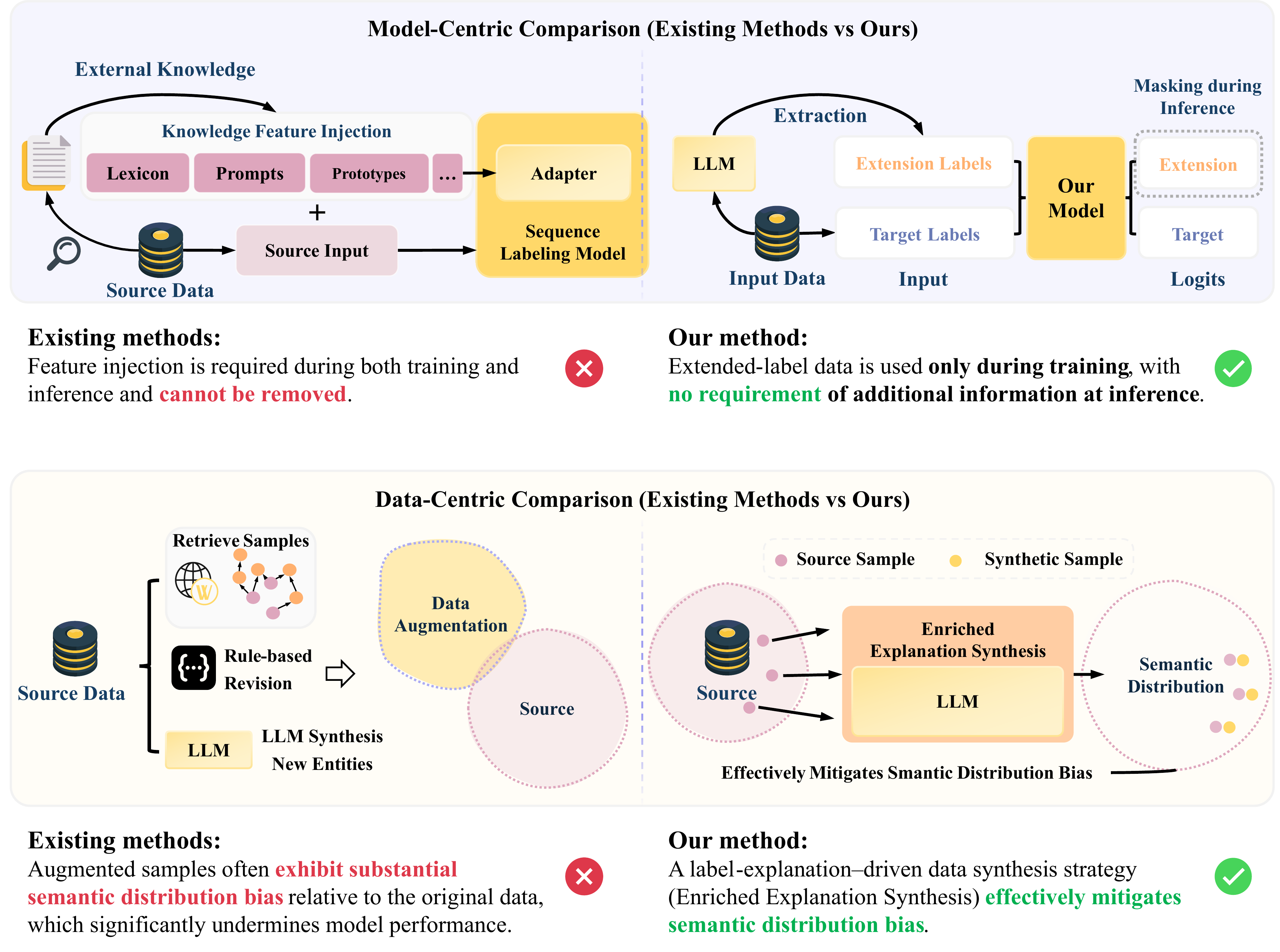}
  \caption{Distinctions between our method and existing methods in terms of model-centric and data-centric.}
  \label{f1}
\end{figure}

Previous studies predominantly focus on two main directions to enhance sequence labeling in low-resource scenarios: \textbf{(1) Model-Centric Optimization.} These methods focus on enhancing the model's comprehension to detect implicit word boundaries and contextual signals through feature engineering. For instance, lexical features are injected via lexicon matching networks \citep{zhang2018chinese, li2020flat, liu2021lexicon, wu-etal-2021-mect} or prompt templates \citep{ma-etal-2022-template, DBLP:conf/acl/0001TWZZX0Z23, DBLP:journals/corr/abs-2109-00720, DBLP:conf/emnlp/DasZS0Z23} to strengthen entity boundary or type detection. Other methods employ knowledge transfer techniques such as Gaussian embeddings \citep{si-etal-2024-improving, DBLP:conf/acl/DasKPZ22}, prompt-based metrics \citep{DBLP:conf/acl/ChenZY23, lai2022pcbert}, and contrastive learning \citep{DBLP:conf/coling/HuangHWZGM022, zhang-etal-2024-kcl} to distill knowledge into target domains. \textbf{(2) Data-Centric Augmentation.} Meanwhile, data-centric methods concentrate on using data augmentation through altering entity label information \citep{hu-etal-2023-entity, yang-etal-2018-distantly}, back translation \citep{DBLP:conf/iclr/PaoliniAKMAASXS21, yaseen-langer-2021-data}, and extracting knowledge from the external environment \citep{cai-etal-2023-graph, chen-etal-2021-data, yaseen-langer-2021-data} to enrich the dataset. With the advent of large language models (LLMs), recent findings leverage their generative capabilities to enhance the diversity of entity and sentence synthesis \citep{DBLP:conf/acl/KangSJJCC24, DBLP:journals/corr/abs-2402-14568}.

However, as illustrated in Figure~\ref{f1}, significant limitations remain when applying these solutions to specialized domains:
\textbf{(1) Limited Model Applicability.} Existing model-centric approaches for character-dense languages often struggle to effectively incorporate diverse feature types and label structures, limiting the flexibility and expressiveness of feature injection. These methods also face difficulties in handling nested entities, further reducing their adaptability. Moreover, many approaches rely on rigid feature integration pipelines and complex input configurations to improve word features, leading to increased reliance on supplementary structures during inference and raising deployment costs. 
\textbf{(2) Variability in Label Distribution.} Existing data-centric augmentation methods frequently suffer from domain distribution biases. Inconsistencies in entity type definitions and semantic contexts across domains lead to mismatches in label priors and entity representations, undermining the quality of synthesized data and weakening zero-shot generalization.

These challenges, including structural rigidity and distributional mismatch, collectively hinder the practical effectiveness of current methods. This motivates our development of a unified framework that addresses both architectural constraints and domain adaptation challenges in a holistic manner. In this task, we adopt two key strategies for improving low-resource sequence labeling in character-dense languages: \textit{(i) enhancing the utilization of non-entity features through the span-based model} and \textit{(ii) improving the model's contextual understanding of target entities.}



 To achieve these objectives, we propose a novel LLM-based data augmentation framework. Our approach begins by designing extraction prompts to identify and extract informative non-target entity features from the input text, thereby maximizing the utilization of non-entity information. \textbf{\textit{To address the issue of limited model applicability,}} we introduce a span-based model called \textbf{Know}ledge \textbf{F}usion for \textbf{R}ich and \textbf{E}fficient \textbf{E}xtraction (\textbf{KnowFREE}), which supports nested entity annotation and integrates extension label features through a local multi-head attention module. Unlike previous methods, KnowFREE captures rich contextual representations during training without relying on external knowledge at inference time. \textbf{\textit{To tackle the issue of variability in label distribution,}} we incorporate explanation prompts inspired by label explanation techniques \citep{golde-etal-2024-large, yang-katiyar-2020-simple, ma-etal-2022-label}, enabling the generation of precise, context-aware explanations for target entities. This enhances the model's contextual understanding and mitigates semantic distribution biases. By leveraging LLMs for label interpretation synthesis, our framework outperforms other related data augmentation techniques in low-resource settings. We evaluate it on multiple Chinese and English domain-specific sequence labeling datasets, and experimental results demonstrate its effectiveness in overcoming the key limitations of low-resource scenarios.

The contributions of our work can be summarized as follows:

(1) \underline{\textit{New method.}} We propose a span-based KnowFREE model that supports nested label annotations and integrates multi-label features by a local multi-head attention module, which can be used without relying on external knowledge at inference.

(2) \underline{\textit{New perspective.}} To the best of our knowledge, we are the first to propose an approach that supports the seamless integration of extension label features within the model while eliminating the need for external features during inference.

(3) \underline{\textit{State-of-the-art performance.}} Experimental results demonstrate that our approach achieves outstanding performance on low-resource sequence labeling tasks.

\section{Related Work}

\textbf{Span-based Sequence Labeling Methods}
Span-based sequence labeling methods have gained prominence for their ability to address overlapping and nested entities effectively \citep{DBLP:conf/acl/YangT22a, DBLP:conf/aaai/FuTCHH21}. Early works, such as \citet{DBLP:conf/iclr/DozatM17, DBLP:conf/acl/YuBP20} introduced Biaffine models to capture sentence-wide structures and score span boundaries for accurate information extraction. Based on this, \citet{DBLP:journals/corr/abs-2208-03054} proposed the Global Pointer model, optimizing the Biaffine transformation's weight matrix and bias terms to boost efficiency and precision in span-based NER. In parallel, \citet{DBLP:conf/acl/0001WTXXH0Z22} introduced a parallel instance query network for simultaneous entity extraction. Then, \citep{DBLP:conf/acl/0001SLQ23} proposed a multi-head Biaffine mechanism combined with CNNs to capture local span features, achieving improved performance, while \citep{DBLP:conf/aaai/Li00WZTJL22} combined bilinear classifiers with dilated convolutions post-CLN to refine span-level relation classification. For few-shot NER, \citep{wang-etal-2022-spanproto} introduced SpanProto, which integrates prototype-based classification with a contrastive loss to effectively separate non-target spans from prototype clusters, demonstrating strong performance in low-resource scenarios.

\textbf{Sequence Labeling via LLMs}

Recent advances in LLMs \citep{openai2023gpt4, deepseekai2025deepseekv3technicalreport, DBLP:journals/corr/abs-2302-13971} have introduced new paradigms for sequence labeling. Generative methods based on in-context learning (ICL) allow LLMs to perform labeling tasks directly without task-specific fine-tuning \citep{DBLP:journals/corr/abs-2405-04960}. In zero-shot settings, InstructUIE \citep{DBLP:journals/corr/abs-2304-08085} adopts a single-turn instruction framework across diverse IE tasks, UniversalNER \citep{DBLP:conf/naacl/0002BLSGI0LLMPR24} demonstrates improved performance by querying one entity type at a time, and GoLLIE \citep{DBLP:conf/iclr/SainzGALRA24} enhances generalization via structured code-style prompting. LLMs have also shown promise in handling cross-domain and nested entity recognition \citep{DBLP:conf/emnlp/NandiA24, DBLP:conf/emnlp/KimKK24}. In parallel, LLM-based data augmentation strategies synthesize high-quality training data by injecting domain-specific features \citep{DBLP:journals/corr/abs-2402-14568, DBLP:conf/acl/HengDLYLZZ24}, while others combine lightweight span detectors with LLM validation to improve span selection in specialized domains \citep{DBLP:conf/emnlp/01560Z0WC24}.


\section{Method}

\begin{figure*}[t]
  \includegraphics[width=\textwidth]{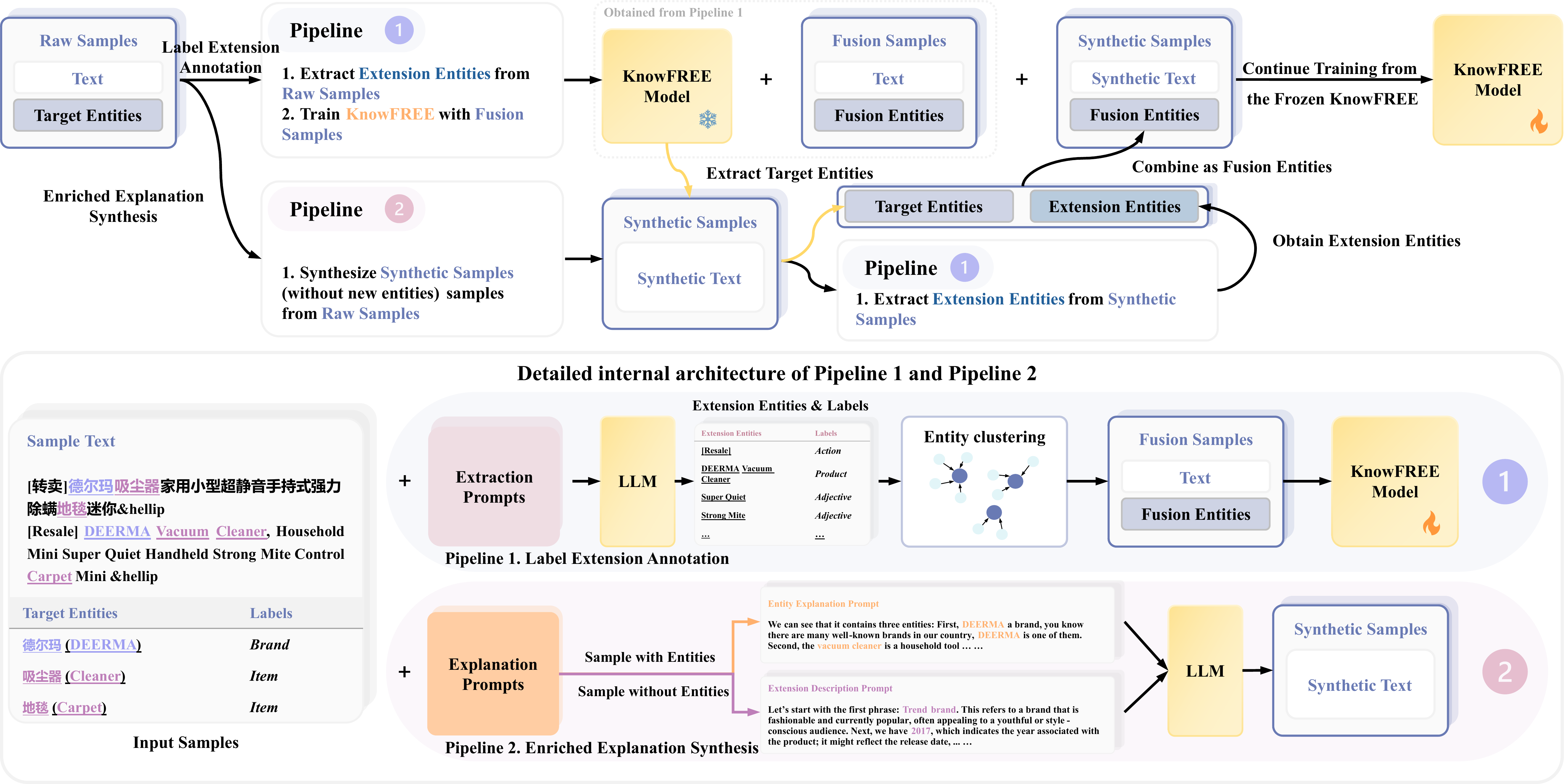}
  \caption{The workflow (top) and detailed pipeline structure (bottom) of our knowledge enhancement framework. Pipeline 1 generates extension entities to enhance the performance of KnowFREE, while Pipeline 2 synthesizes additional training samples and entities. We then use a frozen KnowFREE model to annotate target entities within these synthetic samples.}
  \label{f2}
\end{figure*}

In this section, we will introduce the knowledge enhancement workflow in \S~\ref{s31} and the specific structure of our sequence labeling KnowFREE model in \S~\ref{s32}. The overall framework is shown in Figure~\ref{f2}.

\subsection{Workflow of Knowledge Enhancement}\label{s31}

In the knowledge enhancement workflow, we leverage LLMs to annotate potential entity information in the source sample and provide additional descriptions of entities. This enhances the utilization of non-entity features and improves the model's comprehension of the context in which the target entity appears. Our workflow consists of two main pipelines, which we describe in detail below.

\textbf{Label Extension Annotation.}
In low-resource scenarios, non-entity segments in the sentence may contain additional non-target entity features, while data samples with flat-only entities may potentially contain nested entity information around the target entities. Leveraging this potential feature information can enhance the model's capacity to comprehend the fine-grained semantics and the ability to distinguish entity boundaries in character-dense languages. To achieve this, we utilize the LLM's general knowledge to generate extension entity tags, word segmentation tags, and part-of-speech (POS) tags for the source samples.

Formally, we can denote the source samples as $S = \{s_1, s_2, \dots, s_n\}$, and the LLM as $L$, where $n$ is the number of samples. We then constructed a prompt for each type of tag extraction, denoted as $\mathcal{P}_{ent}$, $\mathcal{P}_{seg}$, and $\mathcal{P}_{pos}$, correspondingly. The extension tags set can be computed as:
\begin{equation}
    \hat{E}^{\left(k\right)} = \bigcup_{i=1}^{n} \hat{E}^{\left(k\right)}_i, \quad \hat{E}^{\left(k\right)}_i = L(s_i, \mathcal{P}_k),
\end{equation}
where $\hat{E}^{\left(k\right)}_i$ represents the extension set using prompt $\mathcal{P}_{k}$ extracted from sentence $s_i$.

We assume that accurate and diverse extension tags can significantly improve the model's comprehension of context and its capacity for entity detection. Then, there are several issues that need to be solved in the results of extraction. (1) The inherent uncertainty in LLM generation often produces extension entity tags in multiple textual forms that nonetheless correspond to the same label category, a phenomenon known as Surface Form Competition \citep{DBLP:conf/emnlp/HoltzmanWSCZ21}. Directly introducing these labels will significantly disrupt the model's assessment of the entity type. (2) Word segmentation usually produces boundary labels with low information entropy. However, POS tagging based on LLMs sometimes has missing annotations. Notably, POS tagging labels inherently include implicit word boundary information. Therefore, combining the outputs of these two processes is expected to yield better results.

To address the first issue, we use LLM to generate synonymous label mapping, and combine entity clustering algorithm to achieve synonymous label merging. Let denote the extension entity set as $\hat{E}^\mathrm{ent}=\{(e_i,t_i)\mid e_i\in\mathcal{E},t_i\in\mathcal{T}\}$, where $\mathcal{E}$ is the set of entities and $\mathcal{T}$ is the set of entity types. We compute the synonymous tag set by using the synonymous tag merge prompt $\mathcal{P}_{merge}$:
\begin{align}
    & \mathcal{T}_s=L(\mathcal{T},\mathcal{P}_{\mathrm{merge}}), \\
    & \mathcal{T}_s=\{\tilde{T}_i:\{T_{i1},T_{i2},\ldots,T_{ir}\}\mid i\in[1,m]\},
\end{align}
where $\tilde{T}_i$ is the standard label, $T_{ij}$ is the synonymous label, and $m$, $r$ is the number of standard label and its corresponding synonymous labels, respectively. This method is determined by the LLM's literal interpretation of the label, which may not accurately align the semantic spatial distribution of entities in the target domain, and therefore is not completely reliable. We further compute the vector representation of each entity-label pair by using a sentence embedding model $\mathcal{M}$, and the vector set of $T_k$ can be represented as:
\begin{align}
    & \mathbf{V}_{k} = \{\mathbf{v}_{i} \mid (e_i, t_i) \in \hat{E}^{\text{ent}}, \mathbf{t}_{i}\in T_{k}\},
\end{align}
where $\mathbf{v}_{i}$ refers to $\mathcal{M}(x_i)$, $x_i$ is the concatenation of $e_i$ and $t_i$ with the template of ``[$e_i$]~is~[$t_i$]''. The center point and mean radius of the vector set for $T_k$ can be calculated as:
\begin{align}
    & \mathbf{c}_k = \frac{1}{|\mathbf{V}_k|} \sum_{\mathbf{v}_i \in \mathbf{V}_k} \mathbf{v}_i, \\
    & r_k = \frac{1}{p} \sum_{j=1}^p \|\mathbf{v}_j - \mathbf{c}_k\|,
\end{align}
where $p$ denotes the Top-$p$ samples that exhibit the greatest distance from $\mathbf{c}_k$. For each standard label $\tilde{T}_i$, we identify the synonymous label vector set $\mathbf{V}_{i,max}$ that contains the largest number of samples and has the largest radius, designating it as the reference vector set. We then evaluate whether each remaining synonymous label vector set $\mathbf{V}_{i,j}$ satisfies the condition $ \|\mathbf{c}_j - \mathbf{c}_{max}\| \leq \epsilon \cdot r_{max}$. If the condition is met, $T_j$ is merged into $\tilde{T}_i$; otherwise, $T_j$ is treated as an independent standard label.

To address the second issue, we first compute the word segmentation set $\hat{E}^\mathrm{seg}$ using $\mathcal{P}_{seg}$. Then, we combine $\hat{E}^\mathrm{seg}$, $\mathcal{P}_{pos}$, and the original source sample as input to the LLM to generate part-of-speech tags without omissions. This approach ensures comprehensive POS tagging for all words in the sample while enhancing the diversity of word segmentation features.

Finally, we merge all the extension entities with standardized labels into the original data to obtain the fusion samples. We use our KnowFREE model to first train an annotation model on the fusion samples, which is then employed to annotate target entities in the synthetic samples generated by the subsequent pipeline.

\begin{figure*}[t!]
  \centering
  \includegraphics[width=0.9\textwidth]{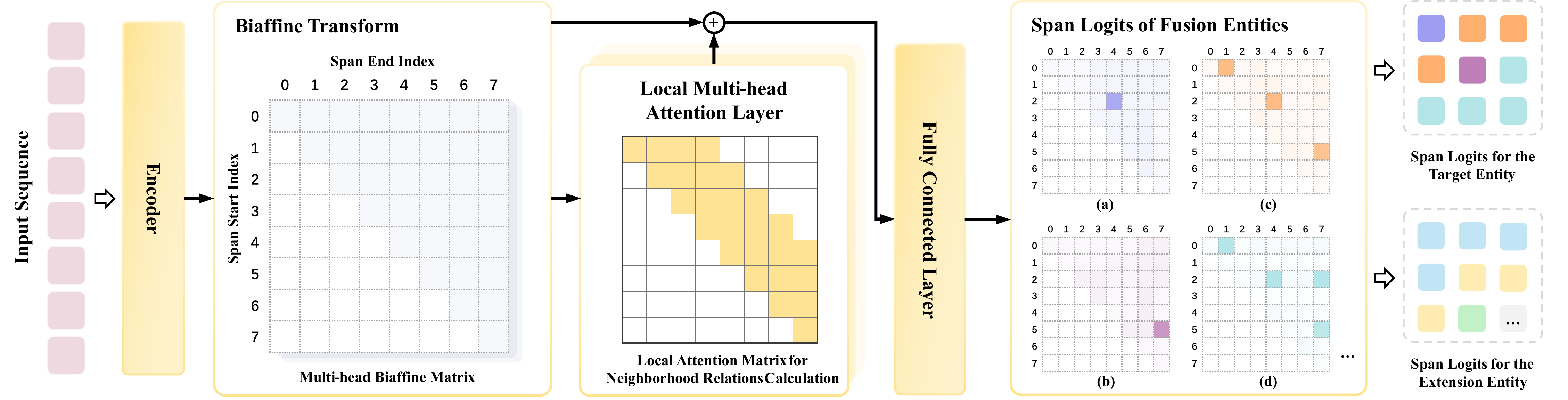}
  \caption{The architecture of the KnowFREE model. The span logits corresponding to the extension entity labels are ignored during inference. Matrices like (a), (b), (c), and (d) represent the span logits for each label type.}
  \label{f3}
\end{figure*}

\textbf{Enriched Explanation Synthesis.}
Injecting extension entity features has been proven to enhance model performance in many cases, its impact remains constrained in the following situations: (1) The datasets with short sentence contains a high proportion of target entities, leading to a relatively low amount of non-entity information in the sample. This results in a diminished validity of the external features introduced. (2) As the quantity of samples diminishes, their scarcity will emerge as the principal factor limiting performance enhancement. Enhancing the exploitation of non-entity features in the sample yields marginal performance enhancement. Both instances highlight the necessity of expanding the sample size. Nonetheless, the divergence in semantic distribution indicates that incorporating samples from outside the source domain, along with directly employing LLM to generate new sentences from existing entities, may introduce potential noise that could significantly affect model performance.

To tackle the aforementioned challenges, we employ LLM to generate entity explanations within the current samples, which aims to leverage the knowledge embedded within the LLM and the connections between samples and entities to produce the precise meaning of target entities within their context. This approach can mitigate noise caused by semantic distribution shifts in synthetic samples. Specifically, we define two types of explanation prompts: the Entity Explanation Prompt $\mathcal{P}_{exp}$ and the Extension Description Prompt $\mathcal{P}_{ext}$ for samples with and without target entities, respectively. The function of $\mathcal{P}_{exp}$ takes the source sample and its corresponding target entity as input, aiming to generate a detailed explanation of the entity's contextual meaning. Meanwhile, $\mathcal{P}_{ext}$ focuses on extracting and explaining key phrases from the text, with only the source sample as input. Additionally, to enrich the semantic representation of the source data, both prompts instruct the LLM to act the role of a domain expert, providing accessible and detailed explanations of the target entities to a hypothetical audience. This strategy encourages the LLM to generate comprehensive, contextually relevant, and easy-to-understand explanations, enhancing the overall semantic clarity of the dataset.

Next, we generate enriched explanations using explanation prompts and annotate entities in the synthetic texts through two branches: one for target entity annotation and the other for extension entity extraction. The frozen KnowFREE model trained in Pipeline 1 is used to annotate target entities, while extension entities are extracted by reapplying Pipeline 1 to the synthetic texts. Finally, the fusion entities of synthetic samples are obtained by integrating both extension and target entities. As shown in Figure \ref{f2}, we combine these fusion samples with synthetic samples and retrain the KnowFREE model to further improve its performance in low-resource scenarios.

\subsection{Structure of KnowFREE Model}\label{s32}

To support the fusion of multi-label knowledge, the KnowFREE model is built upon a Biaffine architecture, as illustrated in Figure~\ref{f3}. Unlike previous methods that rely on external feature injection, our approach eliminates the need for additional injection modules at the input stage.


Formally, let denote $X, \hat{X}$ as inputs of the fusion sample and the synthetic sample, respectively. The target entity spans of each sample is represented as $[s_i, e_i, l_i]$, where $s_i, e_i$ are the start and end indices of the entity span, and $l_i$ is the label type. The extension entity span is represented as $[s_j, e_j, \tilde{l}_j]$, where $\tilde{l}_j$ is the extension label type. We adopt a pre-trained encoder to compute the hidden states $H \in \mathbb{R}^{L \times D}$ for each input, where $L$ is the length of the input sequence, $D$ is the dimension of hidden states. We then compute the encoding of the entity's start and end positions:
\begin{equation}
    H_s = \sigma(H\mathbf{W}_s), \quad H_e = \sigma(H\mathbf{W}_e),
\end{equation}
where $\mathbf{W}_s, \mathbf{W}_e \in \mathbb{R}^{D \times D^\prime}$ are learnable weight matrics, $D^\prime$ is the feature hidden size, and $\sigma$ is the activation function. The Biaffine matrix of spans is computed by a multi-head Biaffine decoder $\mathcal{F}_{\text{MHB}}$ \citep{DBLP:conf/acl/YuBP20}:
\begin{equation}
    H_\text{B} = \mathcal{F}_{\text{MHB}}(H_s, H_e),
\end{equation}
where $H_\text{B} \in \mathbb{R}^{L\times L\times \tilde{D}}$, $\tilde{D}$ is the hidden size of the Biaffine matrix. To improve the interactivity between multi-label features and span neighborhoods, we introduce a local multi-head attention layer to generate a mask for local multi-head attention. Each token is restricted to attend only within a local window of size $\omega$ through a masking scheme:
\begin{equation}
    M[i,j] = 
    \begin{cases} 
    0 & \text{if } |i - j| \leq w, \\
    -\infty & \text{otherwise}.
    \end{cases}
\end{equation}
where $M \in \mathbb{R}^{L \times L}$, $i$, $j$, refer to the row and column indices in $M$. For input features $H_\text{B}$, the attention computation follows the standard multi-head paradigm with $\mathcal{K}$ heads, but incorporates the local mask $M$:
\begin{equation}
\resizebox{0.8\columnwidth}{!}{$ \mathcal{A}(Q,K,V,M) = \text{Softmax}\left(\frac{QK^T}{\sqrt{d_k}} + M\right)V$},
\end{equation}
The outputs $H_\text{attn}$ from all heads are concatenated and processed with layer normalization. We maintain training stability through residual connections:
\begin{equation}
H_\text{G} = \text{LayerNorm}(H_\text{B} + H_\text{attn}).
\end{equation}
We then use the fully connected layer to map the sum of $H_\text{G}$ and $H_\text{B}$ into the number of entity tags:
\begin{equation}
    P = \sigma^{\ast}(\mathbf{W}_{O}(H_\text{B} + H_\text{G}) + \mathbf{b}),
\end{equation}
where $\mathbf{W}_{O} \in \mathbb{R}^{\tilde{D} \times (N_{\text{tgt}} + N_{\text{ext}})}$ is the learnable weight matrix, $\sigma^{\ast}$ is the activation function, and $\mathbf{b} \in \mathbb{R}^{(N_{\text{tgt}} + N_{\text{ext}})}$ is the bias. $N_{\text{tgt}}, N_{\text{ext}}$ are the number of target entity label and extension entity label, respectively. The binary cross entropy is employed to compute the loss. To prevent the model from overly concentrating on features of extension entities, the loss function is defined as:
\begin{equation}
    \resizebox{0.8\columnwidth}{!}{$\mathcal{L} = -(\sum_{\substack{0 \leq i, \\ j < N_{\text{tgt}}}} y_{i,j} \log P_{i,j} + \alpha \sum_{\substack{N_{\text{tgt}} \leq i, \\ j < N_{\text{ext}}}} y_{i,j} \log P_{i,j})$},
\end{equation}
where $y$ refers to the ground truth labels, $\alpha$ is the weight parameter, and $i$, $j$ denote the indices of different label types. Similarly, we control the influence of the quantity and noise in synthetic samples with another weight parameter. The final loss is a weighted sum of the original and synthetic losses:
\begin{equation}
    \mathcal{L}_{\text{final}} = \mathcal{L} + \beta \mathcal{L}_{syn}.
\end{equation}
During inference, the weights associated with the extension entity labels are masked, ensuring that the model exclusively predicts the target entities. This design simplifies the overall architecture while enhancing model efficiency.

\section{Experiment}

\begin{table*}[h!]
\resizebox{\textwidth}{!}{%
\begin{tabular}{@{}l|ccc|ccc|ccc|ccc|ccc|ccc|ccc|ccc@{}}
\toprule
                                                                           & \multicolumn{3}{c|}{\textbf{Weibo}}                       & \multicolumn{3}{c|}{\textbf{Youku}}                       & \multicolumn{3}{c|}{\textbf{Taobao}}                      & \multicolumn{3}{c|}{\textbf{Resume}}                     & \multicolumn{3}{c|}{\textbf{CMeEE-v2}}                    & \multicolumn{3}{c|}{\textbf{PKU}}                   & \multicolumn{3}{c|}{\textbf{MSR}}                        & \multicolumn{3}{c}{\textbf{UD}}                          \\
\multirow{-2}{*}{\textbf{Model}}                                            & 250          & 500          & 1000         & 250          & 500          & 1000         & 250          & 500          & 1000         & 250          & 500          & 1000         & 250          & 500          & 1000         & 250          & 500          & 1000         & 250          & 500          & 1000         & 250          & 500          & 1000         \\ \midrule
BERT-CRF \citep{devlin-etal-2019-bert}                                                                   & 56.57          & 60.91          & 66.52          & 68.02          & 70.57          & 74.92          & 68.78          & 71.88          & 74.74          & 90.19          & 92.35          & 93.43          & -              & -              & -              & 93.49          & 94.31          & 95.02          & 90.60          & 91.73          & 92.93          & 87.11          & 89.87          & 91.98          \\
FLAT \citep{li2020flat}                                                                      & 57.75          & 59.47          & 65.72          & 72.31          & 76.01          & 78.73          & 69.84          & 71.72          & 76.21          & 91.35          & 93.04          & 93.61          & -              & -              & -              & 78.28          & 80.10          & 80.22          & 77.72          & 78.27          & 78.44          & 77.76          & 78.39          & 80.11          \\
MECT \citep{wu-etal-2021-mect}                                                                      & 58.55          & 60.77          & 66.13          & 72.82          & 75.85          & 79.16          & 70.54          & 73.87          & 76.48          & 91.52          & 93.63          & 93.90          & -              & -              & -              & 87.28          & 87.54          & 87.58          & 87.48          & 87.53          & 87.71          & 87.12          & 87.30          & 87.61          \\
LEBERT \citep{liu2021lexicon}                                                                    & 61.23          & 64.03          & 67.63          & 72.39          & 75.00          & 77.88          & 71.12          & 74.46          & 77.44          & 93.08          & 94.16          & 95.05          & -              & -              & -              & 93.42          & 94.09          & 94.97          & 90.64          & 91.95          & 93.27          & 88.93          & 91.85          & 93.43          \\
PCBERT \citep{lai2022pcbert}                                                                    & 70.73          & 70.78          & 72.81          & 77.67          & 81.96          & 83.66          & 73.32          & 75.41          & 79.21          & 93.63          & 94.31          & 95.18          & -              & -              & -              & 93.47          & 94.07          & 94.54          & 90.01          & 92.18          & 93.10          & 89.81          & 91.70          & 93.67          \\ \midrule
BiaffineNER \citep{DBLP:conf/acl/YuBP20}                                                                & 58.74          & 66.41          & 69.70          & 77.28          & 80.21          & 81.68          & 74.62          & 76.98          & 79.46          & 93.30          & 94.61          & 95.49          & 61.26          & 65.56          & 68.40          & 93.20          & 94.39          & 94.94          & 91.60          & 92.63          & 93.64          & 88.84          & 90.74          & 92.68          \\
W$^2$NER \citep{DBLP:conf/aaai/Li00WZTJL22}                                                                      & 54.57          & 63.21          & 71.09          & 79.20          & 81.40          & 83.28          & 74.68          & 76.80          & 79.71          & 94.39          & 95.82          & 96.35          & 61.10          & 65.67          & 68.72          & 93.90          & 94.64          & 95.41          & 91.61          & 92.76          & 93.82          & 90.12          & 92.91          & 94.86          \\
CNN Nested NER \citep{DBLP:conf/acl/0001SLQ23}                                                           & 64.81          & 67.75          & 69.96          & 79.28          & 81.58          & 83.94          & 75.39          & 77.86          & 80.06          & 93.10          & 94.54          & 95.39          & 62.32          & 66.43          & 69.06          & 94.00          & 94.59          & 95.47          & 91.72          & 92.86          & 93.67          & 90.21          & 92.48          & 94.70          \\
DiFiNet \citep{DBLP:conf/acl/Cai0GLLLLJ24}                                                                   & 67.35          & 69.02          & 72.19          & 79.81          & 81.32          & 83.29          & 75.40          & 77.05          & 79.61          & 93.81          & 94.75          & 95.93          & 63.55          & 66.26          & 67.28          & 93.81          & 94.76          & 95.26          & 91.46          & 92.45          & 93.24          & 90.94          & 92.82          & 94.62          \\ \midrule
\rowcolor{gray!10}KnowFREE-F (ChatGLM3-6B)                                                 & 66.76          & 71.59          & 72.99          & 79.30          & 82.13          & 84.50          & 76.31          & \textbf{78.55} & 80.53          & 94.03          & 95.04          & 96.14          & 63.64          & 67.47          & 69.52          & 94.07          & 94.94          & 95.51          & 91.73          & 92.91          & 93.92          & 90.99          & 92.71          & 95.00          \\
\rowcolor{gray!10}KnowFREE-F (GLM4-9B-Chat)                                                       & 66.40          & 73.08          & 72.59          & 79.40          & 82.16          & 84.37          & 76.02          & 78.21          & 80.48          & 94.05          & 95.43          & 96.25          & 63.98          & 67.41          & 69.38          & 94.57          & 95.01          & 95.49          & 91.83          & 92.73          & 93.91          & 90.58          & 92.72          & 94.98          \\
\rowcolor{gray!10}KnowFREE-F (Qwen-14B)                                                      & 68.07          & 73.39          & 73.41          & 80.30          & 82.10          & 84.20          & 76.38          & 78.00          & 80.50          & 94.21          & 95.32          & 96.40          & 63.23          & 67.25          & 69.19          & 94.36          & 94.94          & 95.51          & 91.76          & 92.93          & 93.89          & 90.53          & 93.03          & 94.97          \\
\rowcolor{gray!10}
KnowFREE-F (Llama3.1-70B-Instruct)                                                   & 67.72          & 73.08          & 72.59          & 80.18          & 82.17          & 84.37          & 76.24          & {\ul 78.32}    & 80.51          & 94.11          & 95.24          & 96.18          & 63.92          & 67.47          & 69.18          & 94.28          & 94.98          & 95.51          & 91.77          & 92.89          & 93.91          & 90.76          & 92.69          & 94.96          \\
\rowcolor{gray!10}
KnowFREE-F (GPT-4o)                                                         & 68.18          & 73.51          & 74.06          & 80.30          & 82.18          & {\ul 84.62}    & 76.47          & 78.18          & \textbf{80.64} & 94.55          & {\ul 95.50}    & 96.37          & 63.66          & 67.59          & {\ul 69.76} & 94.59          & {\ul 95.09}    & \textbf{95.62} & 91.97          & 92.96          & \textbf{93.98} & 91.01          & 93.13          & 95.05          \\
\rowcolor{gray!10}
KnowFREE-F (Deepseek-V3) & 68.12          & 73.42          & 74.12          & 80.28          & 82.13          & 84.39          & 76.29          & 78.19          & 80.50          & 94.50          & 95.49          & {\ul 96.42}    & 63.79          & 67.51          & 69.73    & 94.52          & 95.04          & {\ul 95.61}    & 91.79          & 92.97          & 93.88          & 91.10          & 93.30          & 95.12          \\ \midrule
\rowcolor{gray!10} KnowFREE-FS (ChatGLM3-6B) & \textbf{74.78} & {\ul 77.18}    & {\ul 76.78}    & 80.29          & 83.09          & 84.45          & {\ul 76.49}    & 77.94          & 79.54          & 94.71          & 95.40          & 96.18          & 66.80          & {\ul 68.67}    & 69.29          & 94.54          & {\ul 95.09}    & 95.50          & 92.11          & 93.01          & 93.67          & 92.09          & 93.65          & 94.77          \\
\rowcolor{gray!10}KnowFREE-FS (GLM4-9B-Chat)      & {\ul 73.90}    & 76.86          & 76.57          & {\ul 81.86}    & {\ul 83.17}    & 84.48          & 76.47          & 77.89          & 79.36          & {\ul 94.95}    & 95.45          & 96.21          & 68.12          & 68.45          & 68.85          & {\ul 94.73}    & 95.05          & 95.47          & 92.18          & 92.96          & 93.50          & 91.23          & 93.02          & 93.42          \\
\rowcolor{gray!10}KnowFREE-FS (Qwen-14B)     & 73.09          & 73.96          & 74.09          & 81.39          & 82.82          & 83.71          & 76.48          & 77.83          & 79.19          & 94.55          & 95.33          & 96.06          & 66.58          & 67.61          & 68.53          & 94.59          & 95.04          & 95.48          & 92.47          & 93.14          & 93.59          & 92.73          & {\ul 93.88}    & 94.36          \\
\rowcolor[HTML]{F2F2F2} 
\rowcolor{gray!10}KnowFREE-FS (Llama3.1-70B-Instruct) & 74.18          & 76.91          & 76.68          & 81.69          & 83.09          & 84.36          & 76.48          & 77.91          & 79.69          & 94.86          & 95.46          & 96.21          & 68.12          & 68.45          & 68.85          & 94.62          & 95.02          & 95.47          & 92.16          & 93.02          & 93.66          & 92.68          & 93.76          & 94.23          \\
\rowcolor[HTML]{F2F2F2} 
\rowcolor{gray!10}KnowFREE-FS (GPT-4o)       & 74.25          & 77.12          & 77.16          & \textbf{81.98} & 83.16          & 84.48          & 76.47          & 78.12          & 80.02          & 94.59          & 95.49          & 96.31          & {\ul 68.31}    & \textbf{68.73} & \textbf{69.78} & 94.72          & \textbf{95.10} & 95.51          & {\ul 92.62}    & {\ul 93.16}    & {\ul 93.96}    & \textbf{93.72} & \textbf{93.98} & {\ul 95.33}    \\
\rowcolor{gray!10}KnowFREE-FS (Deepseek-V3)  & 74.77          & \textbf{77.19} & \textbf{77.18} & 81.72          & \textbf{83.24} & \textbf{84.97} & \textbf{76.52} & 78.21          & {\ul 80.56}    & \textbf{94.97} & \textbf{95.51} & \textbf{96.46} & \textbf{68.51} & 68.40          & 69.12          & \textbf{94.93} & 95.02          & 95.49          & \textbf{92.83} & \textbf{93.17} & 93.42          & {\ul 93.18}    & 93.81          & \textbf{95.35} \\ \bottomrule
\end{tabular}%
}
\caption{The overall results on many-shot sequence labeling tasks. KnowFREE-F denotes the variant using only the label extension annotation pipeline, while KnowFREE-FS incorporates the enriched explanation synthesis pipeline. The \textbf{bold} values indicate the best performance, and the \underline{underlined} values represent the second-best.}\label{t1}
\end{table*}

\subsection{Experiment Setup}

\textbf{Training:} To comprehensively evaluate the effectiveness of our data augmentation strategy on different LLMs, and to fairly compare previous related methods. We use ChatGLM3-6B, GLM4-9B-Chat, Qwen2.5-14B-Instruct, Llama3.1-70B-Instruct, GPT-4o and Deepseek-V3 \citep{glm2024chatglm, qwen2.5, qwen2, DBLP:journals/corr/abs-2407-21783, openai2023gpt4, deepseekai2025deepseekv3technicalreport} as LLMs for label extension annotation and enriched explanation synthesis. We choose BERT \citep{devlin-etal-2019-bert} as the backbone encoder of the KnowFREE. More detail settings are presented in Appendix~\ref{a_set}. 

\textbf{Evaluation:} To assess the performance of our method in low-resource scenarios, we conducted experiments on a variety of datasets. These include Chinese flat NER datasets (Weibo \citep{DBLP:conf/emnlp/PengD15}, Youku \citep{jie-etal-2019-better}, Taobao \citep{jie-etal-2019-better}, and Resume \citep{DBLP:conf/acl/ZhangY18}); English flat NER datasets (CoNLL'03 \citep{sang2003introduction} and MIT-Movie \citep{liu2013asgard}); a Chinese nested NER dataset (CMeEE-v2 \citep{DBLP:conf/acl/ZhangCBLLSYTXHS22}); word segmentation datasets (PKU and MSR \citep{DBLP:conf/acl-sighan/Emerson05}); and a POS tagging dataset (UD \citep{nivre-etal-2016-universal}). To evaluate our method under data scarcity and explore the limits of performance gains from sample synthesis, we conducted both \textbf{many-shot} and \textbf{few-shot} experiments. In the many-shot setting, we simulated low-resource conditions by randomly sampling subsets of 250, 500, and 1000 training instances. In the few-shot setting, we adopted the standard ``n-way k-shot'' paradigm, using a greedy sampling strategy to ensure each target label appeared at least $k$ times. To ensure consistency, each larger subset included all samples from the smaller ones. Dataset statistics are provided in Appendix~\ref{sta}. We then discuss the effectiveness of each module in the analysis section. Additionally, we report the results on the full datasets in Appendix~\ref{a_full}, conduct further ablation studies on the number of heads in local attention in Appendix~\ref{a_local}, and provide visualizations of the logit scores for the extension labels in Appendix~\ref{a_vis}.

\textbf{Baselines:} To ensure a fair comparison, we evaluate our method against both model-centric and data-centric baselines. On the model-centric side, we compare with general baselines such as BERT-CRF \citep{devlin-etal-2019-bert} and BiaffineNER \citep{DBLP:conf/acl/YuBP20}, as well as models specifically designed for Chinese sequence labeling, including FLAT \citep{li2020flat}, MECT \citep{wu-etal-2021-mect}, and LEBERT \citep{liu2021lexicon}. We also include comparisons with state-of-the-art nested NER methods such as W$^2$NER \citep{DBLP:conf/aaai/Li00WZTJL22}, CNN Nested NER \citep{DBLP:conf/acl/0001SLQ23}, and DiFiNet \citep{DBLP:conf/acl/Cai0GLLLLJ24}. From the data-centric perspective, we compare with the transfer learning approach PCBERT \citep{lai2022pcbert}, and LLM-enhanced methods including LLM-DA \citep{DBLP:journals/corr/abs-2402-14568}, ProgGen \citep{DBLP:conf/acl/HengDLYLZZ24}, and MELM \citep{DBLP:conf/acl/ZhouLHBCSM22}. In addition, Appendix~\ref{a_llms} provides results comparing vanilla LLMs and LoRA fine-tuned models on sequence labeling tasks.

\subsection{Main Results}

\textbf{Many-shot Results.} As shown in Table~\ref{t1}, our method consistently achieves higher average performance across all datasets. We observe that larger backbone LLMs generally bring greater performance improvements to our approach. Although the scaling law is not strictly linear, even the smallest model, ChatGLM3-6B, delivers strong results. Under the 250-sample setting, our method surpasses the strongest baseline by an average of 1.95\%, especially with a 4.05\% gain on the Weibo dataset. In low-resource settings, KnowFREE-FS outperforms KnowFREE-F. However, as the number of training samples increases, especially beyond 500, the performance of KnowFREE-FS becomes comparable to or slightly lower than that of KnowFREE-F. This indicates that enriched data synthesis is more effective when training data is limited. When more data is available, the noise introduced by synthetic samples may outweigh the benefits. Further analysis of noise effects is provided in Appendix~\ref{a_da}. Even when data synthesis becomes less effective in higher-resource scenarios, KnowFREE-F maintains strong performance. With 1000 training samples, it still outperforms the strongest baseline by an average of 0.95\% across all NER datasets, demonstrating the robustness and effectiveness of the label extension annotation strategy.

\begin{table}[t!]
\resizebox{\columnwidth}{!}{%
\begin{tabular}{@{}l|cccccc@{}}
\toprule
\multirow{2}{*}{Size} & \textbf{Ours} & LLM-DA & ProgGen & MELM  & C.N.-NER & DiFiNet \\
                       & \multicolumn{6}{c}{Weibo}                                      \\ \midrule
$k$=5                    & \textbf{35.58 (2.33)} & 0.00 (0.00)   & 0.00 (0.00)    & 0.00 (0.00)  & 0.00 (0.00)           & 0.00 (0.00)    \\
$k$=10                   & \textbf{48.53 (1.57)} & 0.95 (0.23)   & 0.48 (0.19)    & 0.03 (0.04)  & 0.95 (0.21)           & 0.98 (0.18)    \\
$k$=15                   & \textbf{60.61 (2.58)} & 17.28 (1.93)  & 12.36 (1.22)   & 2.98 (0.58)  & 20.68 (1.88)          & 27.43 (2.30)   \\
$k$=20                   & \textbf{68.62 (2.04)} & 39.49 (2.41)  & 34.16 (2.09)   & 16.71 (1.67) & 31.10 (2.01)          & 32.71 (2.31)   \\ \midrule
                       & \multicolumn{6}{c}{Youku}                                            \\ \midrule
$k$=5                    & \textbf{38.76 (2.30)} & 12.03 (2.24)  & 13.76 (1.32)   & 9.98 (2.01)  & 24.70 (2.55)          & 23.83 (2.52)   \\
$k$=10                   & \textbf{68.95 (3.85)} & 33.10 (2.34)  & 33.52 (2.01)   & 16.17 (1.30) & 46.39 (2.48)          & 46.62 (2.43)   \\
$k$=15                   & \textbf{71.76 (3.33)} & 59.72 (3.94)  & 56.55 (1.56)   & 50.18 (1.67) & 60.41 (1.88)          & 60.61 (1.91)   \\
$k$=20                   & \textbf{72.83 (2.29)} & 64.58 (2.53)  & 61.80 (1.60)   & 52.38 (1.59) & 67.38 (1.92)          & 68.38 (1.58)   \\ \midrule
                       & \multicolumn{6}{c}{Taobao}                                           \\ \midrule
$k$=5                    & \textbf{62.96 (2.34)} & 13.95 (2.32)  & 22.74 (2.62)   & 8.97 (0.88)  & 22.40 (1.61)          & 23.82 (1.56)   \\
$k$=10                   & \textbf{64.64 (3.58)} & 54.05 (2.92)  & 50.75 (1.66)   & 32.41 (1.37) & 53.33 (2.01)          & 53.75 (2.03)   \\
$k$=15                   & \textbf{69.26 (2.65)} & 59.69 (2.58)  & 59.77 (1.51)   & 55.32 (1.57) & 60.45 (1.83)          & 61.01 (1.54)   \\
$k$=20                   & \textbf{68.93 (2.36)} & 61.79 (1.51)  & 61.77 (1.63)   & 43.13 (1.62) & 63.36 (1.86)          & 64.08 (1.53)   \\ \midrule
                       & \multicolumn{6}{c}{Resume}                                           \\ \midrule
$k$=5                    & \textbf{65.28 (0.97)} & 30.44 (0.66)  & 27.63 (0.34)   & 20.67 (1.26) & 25.50 (1.22)          & 29.97 (1.17)   \\
$k$=10                   & \textbf{78.89 (0.53)} & 45.40 (0.58)  & 50.39 (0.82)   & 42.34 (1.28) & 49.78 (1.53)          & 50.19 (1.49)   \\
$k$=15                   & \textbf{85.43 (1.17)} & 58.77 (1.15)  & 60.32 (1.26)   & 53.18 (1.18) & 56.04 (1.31)          & 56.92 (1.16)   \\
$k$=20                   & \textbf{85.56 (1.11)} & 75.13 (1.36)  & 82.96 (1.28)   & 69.15 (1.21) & 67.51 (1.17)          & 68.83 (1.19)   \\ \midrule
                       & \multicolumn{6}{c}{CMeEE-v2}                                         \\ \midrule
$k$=5                    & \textbf{49.68 (1.89)} & 35.03 (1.65)  & 39.18 (1.52)   & 12.78 (1.01) & 6.49 (0.56)           & 5.62 (0.47)    \\
$k$=10                   & \textbf{60.46 (1.67)} & 48.91 (1.61)  & 44.80 (1.59)   & 32.76 (1.65) & 47.40 (1.56)          & 47.25 (1.50)   \\
$k$=15                   & \textbf{62.46 (1.51)} & 48.97 (1.54)  & 49.32 (1.61)   & 38.79 (1.53) & 48.90 (1.65)          & 48.72 (1.59)   \\
$k$=20                   & \textbf{63.83 (1.68)} & 57.18 (1.63)  & 57.76 (1.51)   & 50.12 (1.67) & 56.38 (1.58)          & 56.22 (1.55)   \\ \midrule
                       & \multicolumn{6}{c}{CoNLL'03}                                         \\ \midrule
$k$=5                    & \textbf{64.18 (1.62)} & 57.84 (1.93)  & 57.11 (2.56)   & 30.00 (2.13) & 27.75 (1.61)          & 26.86 (1.36)   \\
$k$=10                   & \textbf{75.83 (1.52)} & 69.07 (2.44)  & 69.10 (2.39)   & 63.48 (2.18) & 62.93 (1.94)          & 59.17 (1.88)   \\
$k$=15                   & \textbf{78.68 (1.39)} & 78.18 (2.22)  & 78.32 (2.19)   & 76.16 (2.03) & 75.93 (1.86)          & 75.85 (1.89)   \\
$k$=20                   & \textbf{83.24 (1.21)} & 81.94 (2.17)  & 82.09 (1.55)   & 79.41 (2.21) & 77.92 (1.58)          & 77.74 (1.81)   \\ \midrule
                       & \multicolumn{6}{c}{MIT-Movie}                                        \\ \midrule
$k$=5                    & \textbf{57.34 (1.88)} & 53.97 (2.07)  & 52.82 (2.44)   & 38.49 (2.23) & 36.81 (1.26)          & 37.62 (1.33)   \\
$k$=10                   & \textbf{64.08 (1.66)} & 63.03 (2.35)  & 63.41 (2.14)   & 50.56 (2.29) & 49.08 (1.60)          & 49.43 (1.58)   \\
$k$=15                   & \textbf{67.03 (1.68)} & 65.77 (1.46)  & 65.93 (1.17)   & 58.32 (1.87) & 58.03 (1.69)          & 58.54 (1.51)   \\
$k$=20                   & \textbf{69.28 (1.63)} & 69.12 (1.08)  & 69.19 (1.59)   & 62.02 (1.91) & 60.81 (1.60)          & 61.07 (1.64)   \\ \bottomrule
\end{tabular}%
}
\caption{Results of few-shot sequence labeling tasks. Our default method is KnowFREE-FS (ChatGLM3-6B). C.N.-NER refers to the abbreviation of CNN Nested NER. Values in parentheses indicate standard deviation. \textbf{bold} numbers highlight the best performance.}
\label{t8}
\end{table}

\textbf{Few-shot Results.} To assess the effectiveness of our method in few-shot settings, we compared it with state-of-the-art nested NER models and several LLM-based data augmentation strategies, including LLM-DA, ProgGen, and MELM, on both Chinese and English NER datasets. For fair comparison, all synthesized samples were annotated using the KnowFREE model trained solely on the original data, and the resulting data were used to retrain KnowFREE. As shown in Table~\ref{t8}, our method consistently outperforms the baselines under few-shot settings. On the Weibo dataset with $k$=5, while other methods yield zero performance, our approach achieves the performance of 35.58\%. Moreover, for $k \leq 15$, LLM-based augmentation strategies often perform worse than CNN Nested NER and DiFiNet, indicating limited domain adaptability and the adverse effects of noise introduced by data synthesis. The performance gains are more pronounced on Chinese datasets compared to English ones, demonstrating the method's robustness across languages and its particular strength in character-dense languages. Further analysis of performance trends of LLM-based methods under varying data sizes is provided in Appendix~\ref{a_da}.

\subsection{Analysis}

\begin{table}[h!]
  \centering
  \resizebox{\columnwidth}{!}{%
  \begin{tabular}{lcccccccc}
      \toprule \textbf{Method} & \textbf{Weibo} & \textbf{Youku} & \textbf{Taobao} & \textbf{Resume} & \textbf{CMeEE-v2} & \textbf{PKU} & \textbf{MSR} & \textbf{UD} \\
      \midrule Default         & \textbf{72.99}          & \textbf{84.50}          & \textbf{80.53}           & \textbf{96.14}           & \textbf{69.52}             & \textbf{95.51}        & 93.92        & 95.00       \\
      \hspace{1em}w/o L.E.A.     & 72.32          & 84.26          & 79.93           & 96.02           & 69.49             & 93.75        & 93.67        & 94.88       \\
      \hspace{1em}w/o local attn \& L.E.A. & 69.87 & 81.65 & 79.01  & 95.49  & 68.42 & 94.93 & 93.64 & 92.66 \\
      \hspace{1em}w/o local attn w cnn    & 72.21 & 84.28 & 80.26  & 95.94  & 69.31 & 95.50 & 93.72 & 94.76 \\
      \hspace{1em}w/o S.L.     & 72.26          & 84.19          & 79.80           & 95.74           & 69.18             & 94.86        & 93.87        & 94.85       \\
      \hspace{1em}w/o entity   & 72.25          & 84.37          & 80.14           & 95.86           & 69.08             & 95.51        & 93.91        & 94.97       \\
      \hspace{1em}w/o pos      & 72.39          & 84.45          & 80.48           & \textbf{96.14}           & 69.26             & \textbf{95.51}        & \textbf{93.93}        & \textbf{95.01}       \\
      \bottomrule
  \end{tabular}%
  }
  \caption{Results of F1 scores in ablation studies, all results are trained on datasets with 1000. The backbone LLM is ChatGLM3-6B. }
  \label{t3}
\end{table}

\textbf{Ablation Studies:} To evaluate the contribution of each component in our approach, we conducted ablation studies by selectively removing modules and analyzing their impact on model performance, as shown in Table~\ref{t3}. ``w/o L.E.A.'' removes the Label Extension Annotation module and uses the vanilla KnowFree model. Although this leads to a performance drop, it still outperforms nested NER baselines across several datasets. ``w/o local attn \& L.E.A.'' disables both the local attention and L.E.A. modules, resulting in a significant average performance drop of 1.56\%, highlighting their combined effectiveness. In ``w/o local attn w CNN,'' the local attention module is replaced by the masked CNN module from CNN Nested NER. While this improves performance over CNN Nested NER, it underperforms compared to our attention-based model, confirming the advantage of local multi-head attention for capturing neighborhood interactions. ``w/o S.L.'' removes the synonymous label merging strategy and causes a 0.42\% drop in performance, indicating that failing to unify semantically equivalent labels introduces confusion and weakens model predictions. In ``w/o entity'' and ``w/o pos,'' we exclude extension entity features and POS features, respectively. Removing entity features leads to a larger performance drop, showing their stronger impact on entity recognition. Interestingly, removing POS features improves results on MSR and UD, possibly due to noise introduced by imperfect or overly correlated POS tags.

\begin{figure}[t!]
  \centering
  \includegraphics[width=0.9\columnwidth]{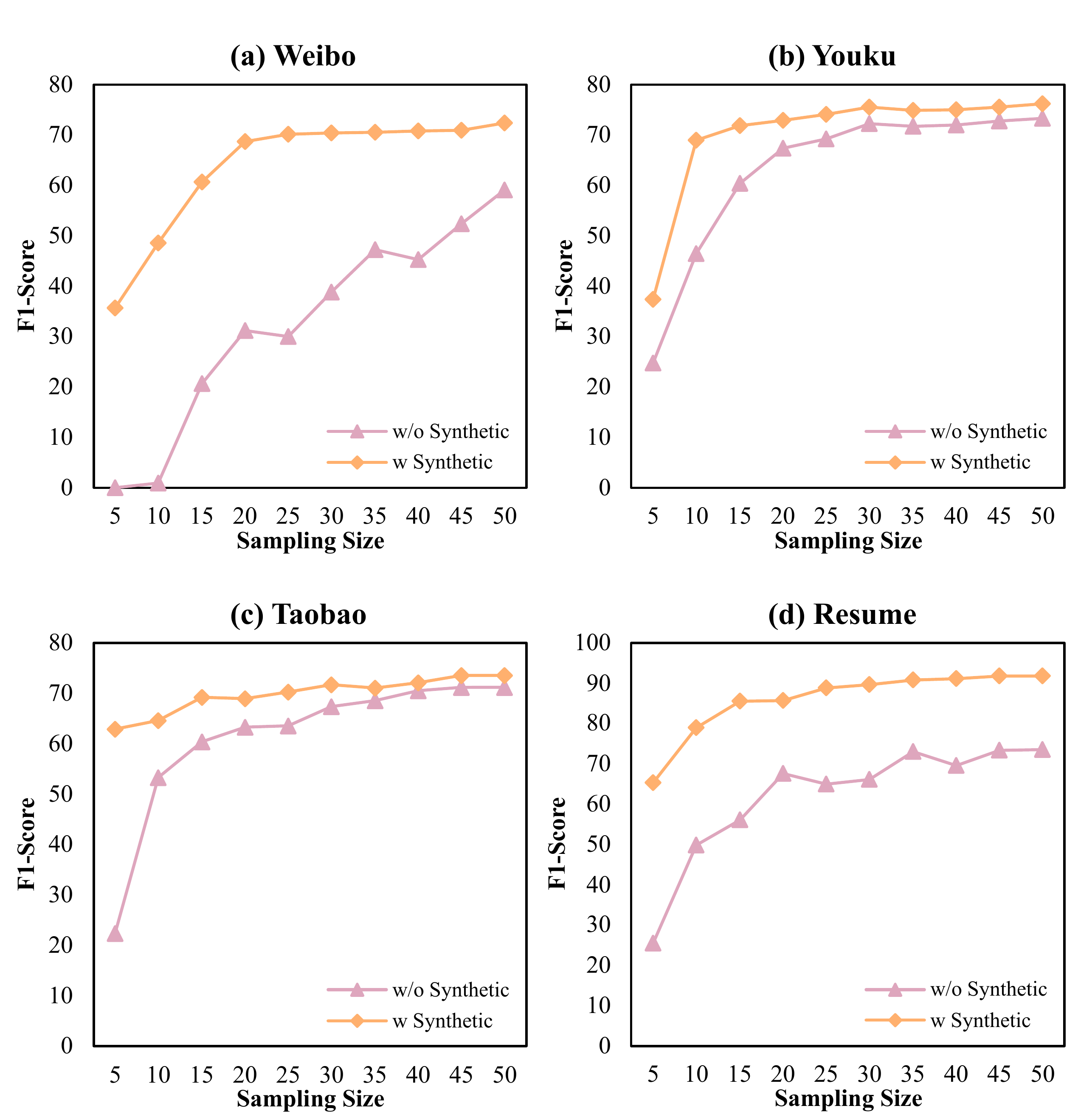}
  \caption{Performance comparison with and without enriched explanation synthesis under k-shot sampling.}
  \label{f4}
\end{figure}

\begin{figure*}[h]
  \includegraphics[width=\textwidth]{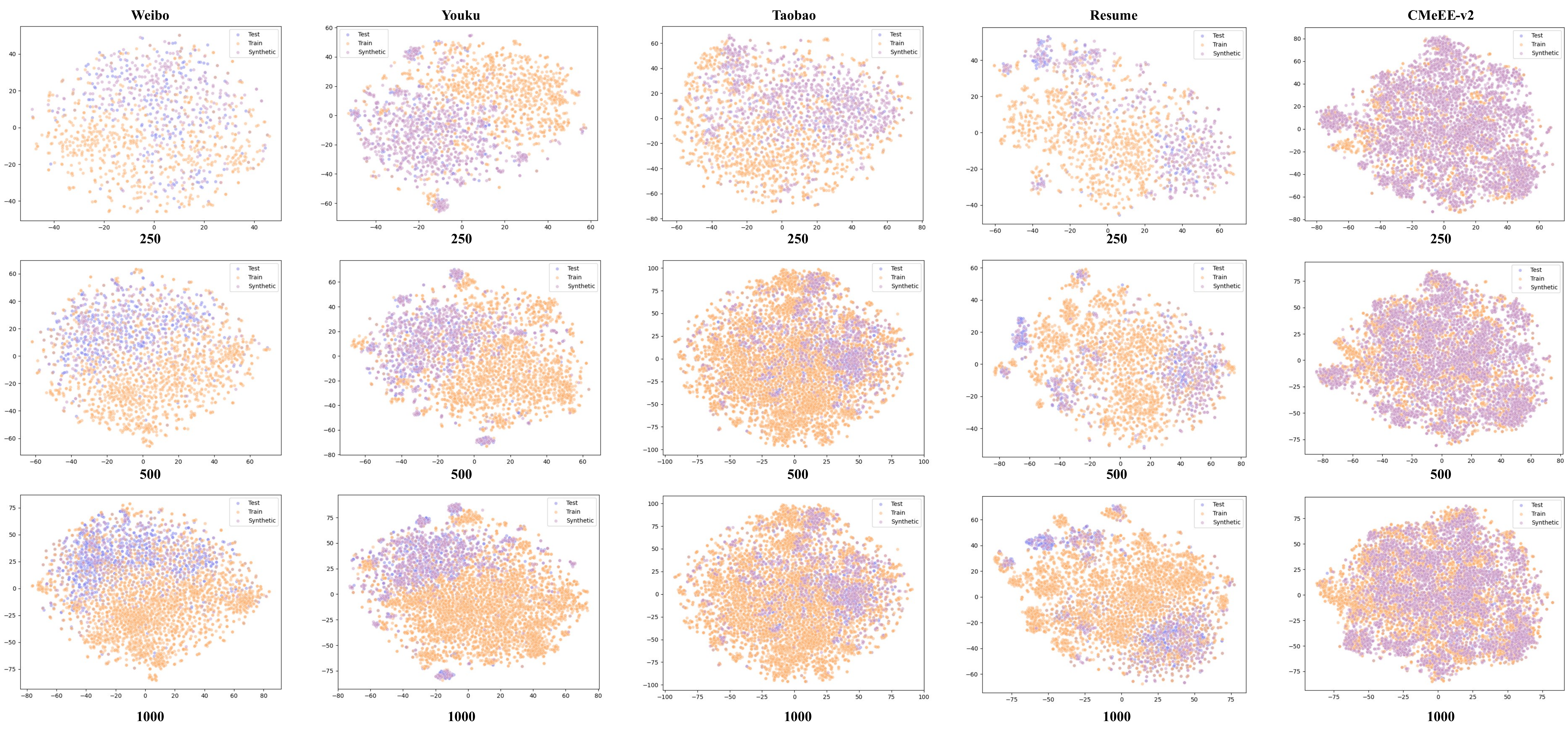}
  \caption{t-SNE visualization of the training, test and enriched explanation samples under different sampling sizes. The synthetic enriched explanation samples are generated by ChatGLM3-6B, and they are represented by the ``Synthetic'' in the legend.}
  \label{f5}
\end{figure*}

\textbf{Impact of Enriched Explanation Synthesis in K-Shot Sampling:} To further evaluate the impact of enriched explanation synthesis on model performance in low-resource scenarios, we conducted experiments following the ``n-way k-shot'' paradigm. The sampled data was then augmented with ChatGLM3-6B. We compared the model's performance with and without enriched explanation synthesis, as shown in Figure~\ref{f4}. Here, ``w Synthetic'' indicates the performance with enriched explanation synthesis, while ``w/o Synthetic'' reflects the performance without it. The results demonstrate that ``w Synthetic'' achieves a substantial performance boost over ``w/o Synthetic'' from the outset. Notably, when k is less than 15, enriched explanation synthesis consistently delivers rapid performance improvements across all datasets. As k increases, the performance gap narrows, but ``w Synthetic'' continues to outperform ``w/o Synthetic'' across all settings. These findings highlight that in resource-scarce scenarios, synthesizing enriched data is more effective than directly injecting features into raw samples. These findings highlight the critical role of enriched explanation synthesis in enhancing model performance, particularly when labeled data is limited.

\textbf{Visual Analysis of Enriched Explanation Synthesis:} We use the text2vec embedding model \citep{text2vec} to generate sentence embeddings for the original training and test samples, as well as for the enriched explanation samples from the Weibo, Youku, Taobao, Resume, and CMeEE-v2 datasets. These embeddings are projected into two dimensions using t-SNE, with results shown in Figure~\ref{f5}. At a sample size of 250, the training data in datasets such as Weibo, Youku, Taobao, and Resume provides sparse semantic coverage, leaving portions of the test set insufficiently represented. This limitation is observed across all datasets. The synthesized samples help bridge these gaps in semantic space, which explains the substantial performance gains in low-resource scenarios. As the number of training samples increases, the coverage of the semantic space becomes more comprehensive for most datasets. However, at a sample size of 1000, semantic discrepancies appear between training and synthesized samples in datasets such as Youku, Taobao, and CMeEE-v2, which may introduce noise and hinder model performance. In the Weibo dataset, certain regions of the test semantic space remain underrepresented even at this larger sample size. This observation explains why models trained with synthesized samples continue to outperform those trained solely on original data at this sample size.

These findings underscore the effectiveness of enriched explanation synthesis in enhancing model performance under low-resource conditions. However, they also reveal that as the sample size grows, the potential drawbacks of synthesized data, such as semantic noise, become increasingly evident.

\section{Conclusion}
In this paper, we propose a novel framework that integrates an LLM-based knowledge enhancement workflow with a span-based sequence labeling model. Our approach improves model performance by generating contextual interpretations of target entities and annotating extension labels. Additionally, our KnowFREE model effectively incorporates extension label features to enhance extraction capabilities. Extensive experiments demonstrate that our method achieves state-of-the-art performance, showcasing its effectiveness and efficiency.

\section*{Limitations}
While enriched explanation synthesis significantly improves model performance in low-resource scenarios (e.g., with fewer than 500 original samples), its effectiveness diminishes as the size of the original dataset increases. Specifically, when the number of original samples exceeds this threshold, distributional discrepancies between synthetic samples and target domain semantics can lead to the synthetic data having a negative impact that outweighs its benefits. In future work, we plan to explore adaptive alignment mechanisms to better align synthetic and original data across different data scales.

\section*{Ethics Statement}
Our data augmentation method utilizes LLMs to generate data independently of the existing training set. However, the generated data may reflect social biases inherent in the pre-training corpus. To mitigate the risk of propagating biased information into sequence labeling models, we recommend conducting manual reviews before integrating the synthesized data into practical applications.

\section*{Acknowledgments}
This work is supported by National Natural Science Foundation of China (U23B2048, 62402011) and High-performance Computing Platform of Peking University. Bin Cui is the corresponding author.


\bibliography{custom}

\appendix

\section{Sequence Labeling with LLMs}\label{a_llms}

\begin{table}[h]
\centering
\resizebox{\columnwidth}{!}{%
\begin{tabular}{lcccccccc}
\toprule
\textbf{Model} & \textbf{Weibo} & \textbf{Youku} & \textbf{Taobao} & \textbf{Resume} & \textbf{CMeEE} & \textbf{UD} & \textbf{PKU} & \textbf{MSR} \\ \midrule
ChatGLM3-6B     & 3.28           & 9.84           & 9.21           & 13.71          & 2.41           & 1.15           & 3.76           & 6.69           \\
Llama3-8B-Instruct     & 15.06          & 13.89          & 9.67           & 41.45          & 11.27          & 14.06          & 23.88          & 28.84          \\
GLM4-9B-Chat     & 12.43          & 31.69          & 14.90          & 47.12          & 16.28          & 26.38          & \textbf{26.18} & \textbf{31.12} \\
Qwen2.5-14B-Instruct & 31.06          & \textbf{51.00} & 5.89           & 51.67          & 22.38          & 10.06          & 2.33           & 3.02           \\
Llama3.1-70B-Instruct  & \textbf{34.69} & 49.54          & 13.77          & \textbf{66.60} & 40.86          & 37.82          & 3.96           & 7.44           \\
Deepseek-V3   & 33.33          & 10.91          & \textbf{15.91} & 60.88          & \textbf{42.20} & \textbf{38.66} & 5.96           & 2.45           \\ \bottomrule
\end{tabular}%
}
\caption{Results on sequence labeling tasks with vanilla LLMs on zero-shot learning.}\label{t5}
\end{table}

\begin{table*}[h]
    \centering
    \resizebox{\textwidth}{!}{%
    \begin{tabular}{l|ccc|ccc|ccc|ccc}
        \toprule \multirow{2}{*}{\textbf{Model}} & \multicolumn{3}{c|}{\textbf{Weibo}}    & \multicolumn{3}{c|}{\textbf{Youku}} & \multicolumn{3}{c|}{\textbf{Taobao}} & \multicolumn{3}{c}{\textbf{Resume}} \\
                                             & 250                      & 500                   & 1000                   & 250                  & 500 & 1000 & 250 & 500 & 1000 & 250 & 500 & 1000 \\
        \midrule ChatGLM3-6B (LoRA)                          & 12.29                                   & 20.44                                & 28.22                                 & 34.52                              & 38.62             & 39.29              & 29.91              & 30.02              & 30.03               & 39.92             & 40.76             & 39.98              \\
        GLM4-9B-Chat (LoRA)                                   & 49.20                                  & 54.54                               & \textbf{60.30}                       & \textbf{73.79}                     & 74.56             & 74.84              & 58.84             & \textbf{63.63}    & \textbf{69.20}     & 83.31             & 86.24             & 88.27              \\
        Llama3-8B-Instruct (LoRA)                & 10.09   & 17.08   & 44.24    & 49.59  & 60.97  & 64.33   & 23.84   & 34.65  & 42.98   & 65.36   & 70.82  & 74.86   \\
        Qwen2.5-14B-Instruct (LoRA)                               & \textbf{50.93}                         & \textbf{56.87}                      & 56.30                                & 72.89                              & \textbf{75.10}    & \textbf{78.83}     & \textbf{61.58}    & 59.43             & 66.98              & \textbf{84.45}    & \textbf{88.12}    & \textbf{88.65}     \\
        \midrule \multirow{2}{*}{\textbf{Model}} & \multicolumn{3}{c|}{\textbf{CMeEE-v2}} & \multicolumn{3}{c|}{\textbf{PKU}}   & \multicolumn{3}{c|}{\textbf{MSR}}    & \multicolumn{3}{c}{\textbf{UD}}     \\
                                             & 250                      & 500                   & 1000                   & 250                  & 500 & 1000 & 250 & 500 & 1000 & 250 & 500 & 1000 \\
        \midrule ChatGLM3-6B (LoRA)                          & 12.75                                   & 13.30                                & 13.55                                 & 20.62                               & 26.27              & 29.21               & 23.53              & 26.04              & 29.38               & 32.20              & 37.57              & 37.81               \\
        GLM4-9B-Chat (LoRA)                                   & 50.67                                  & \textbf{54.50}                      & 56.22                                & 71.41                              & \textbf{73.68}    & 75.02              & 75.70             & 78.64             & \textbf{80.94}     & 72.32             & \textbf{77.50}    & 79.69              \\
        Llama3-8B-Instruct (LoRA)                & 23.27   & 41.52   & 44.41    & 62.50  & 64.53  & 68.88   & 66.17   & 68.52  & 69.72   & 32.16   & 35.51  & 44.85   \\
        Qwen2.5-14B-Instruct (LoRA)                               & \textbf{53.59}                         & 53.80                               & \textbf{59.31}                       & \textbf{71.94}                     & 72.71             & \textbf{75.91}     & \textbf{77.83}    & \textbf{79.89}    & 80.07              & \textbf{74.40}    & 77.04             & \textbf{80.70}     \\
        \bottomrule
    \end{tabular}%
    }
    \caption{Performance of LLM fine-tuning with LoRA on sequence labeling tasks with source datasets.}
    \label{t6}
\end{table*}

\begin{table*}[h]
    \centering
    \resizebox{\textwidth}{!}{%
    \begin{tabular}{l|ccc|ccc|ccc|ccc}
        \toprule \multirow{2}{*}{\textbf{Model}} & \multicolumn{3}{c|}{\textbf{Weibo}}    & \multicolumn{3}{c|}{\textbf{Youku}} & \multicolumn{3}{c|}{\textbf{Taobao}} & \multicolumn{3}{c}{\textbf{Resume}} \\
                                             & 250                      & 500                   & 1000                   & 250                  & 500 & 1000 & 250 & 500 & 1000 & 250 & 500 & 1000 \\
        \midrule 
        ChatGLM3-6B (LoRA)          & 5.81           & 6.11           & 6.09           & 11.98          & 12.33          & 12.37          & 3.69           & 3.92           & 3.98           & 10.88          & 12.69          & 13.98          \\
GLM4-9B-Chat (LoRA)         & 40.22          & \textbf{58.18} & \textbf{62.38} & \textbf{69.81} & \textbf{75.72} & \textbf{74.68} & 59.03          & \textbf{63.22} & \textbf{69.03} & \textbf{83.47} & 86.34          & 88.10          \\
Llama3-8B-Instruct (LoRA)   & 22.08          & 12.42          & 11.79          & 18.97          & 19.85          & 20.35          & 20.22          & 16.76          & 16.12          & 49.10          & 46.11          & 45.65          \\
Qwen2.5-14B-Instruct (LoRA) & \textbf{46.22} & 51.20          & 56.06          & 31.92          & 19.40          & 20.60          & \textbf{62.25} & 62.26          & 65.10          & 82.50          & \textbf{87.28} & \textbf{88.54} \\ \midrule
        \multirow{2}{*}{\textbf{Model}} & \multicolumn{3}{c|}{\textbf{CMeEE-v2}} & \multicolumn{3}{c|}{\textbf{PKU}}   & \multicolumn{3}{c|}{\textbf{MSR}}    & \multicolumn{3}{c}{\textbf{UD}}     \\
                                             & 250                      & 500                   & 1000                   & 250                  & 500 & 1000 & 250 & 500 & 1000 & 250 & 500 & 1000 \\
        \midrule
        ChatGLM3-6B (LoRA)          & 1.14           & 1.02           & 1.16           & 1.89           & 1.78           & 1.69           & 4.71           & 7.62           & 7.75           & 5.65           & 6.17           & 6.68           \\
GLM4-9B-Chat (LoRA)         & 50.02          & \textbf{54.42} & 56.79          & 70.57          & \textbf{73.18} & 75.01          & 75.33          & 78.28          & 80.26          & 70.94          & \textbf{78.41} & 79.08          \\
Llama3-8B-Instruct (LoRA)   & 20.69          & 19.86          & 19.29          & 10.56          & 9.48           & 8.69           & 27.36          & 26.37          & 25.64          & 31.25          & 30.18          & 30.02          \\
Qwen2.5-14B-Instruct (LoRA) & \textbf{52.17} & 54.19          & \textbf{60.83} & \textbf{72.01} & 72.81          & \textbf{75.55} & \textbf{77.48} & \textbf{79.22} & \textbf{80.95} & \textbf{74.09} & 77.41          & \textbf{80.55} \\ \bottomrule
    \end{tabular}%
    }
    \caption{Performance of LLM fine-tuning with LoRA on sequence labeling tasks augmented with synthetic datasets.}
    \label{t10}
\end{table*}

To evaluate the performance of directly using LLMs for sequence labeling tasks in low-resource scenarios, we present the results of vanilla LLMs in Table~\ref{t5} and the performance of LoRA-finetuned models trained on sampled datasets in Table~\ref{t6}. During zero-shot inference, LLMs extract entities based on the input text and target labels. As shown in Table~\ref{t5}, vanilla LLMs exhibit significantly poor performance on all datasets, likely due to their limited understanding of target label definitions. Moreover, models with different parameter scales show varying performance across datasets, and no clear positive correlation is observed between model size and performance. This may be improved through more advanced designs of the sequence labeling prompts used in our setting. Therefore, for domain-specific sequence labeling tasks, incorporating few-shot examples into the prompt or fine-tuning the model with LoRA could be a more effective and practical approach.

In the experiments with LoRA fine-tuning, all LLMs show performance improvements compared to zero-shot inference. Compared to other LLMs, ChatGLM3-6B still underperforms in LoRA fine-tuning, likely due to its weaker ability to align with the target domain of sequence labeling. As a result, the synthesized data produced by ChatGLM3-6B contains a considerable amount of irrelevant information. With the 250-sample setting, Qwen2.5-14B-Instruct demonstrates outperforms other LLMs on most datasets, suggesting that models with larger parameters tend to show enhanced initial performance in low-resource contexts. Nonetheless, with the sample size increases, the performance improvements of Qwen2.5-14B-Instruct on datasets such as Weibo, Taobao, and MSR fell short compared to GLM4-9B-Chat. This could be due to variations in model performance when applied to different domain-specific data distributions. It is important to highlight that the performance of LLMs on most datasets was still below that of conventional sequence labeling baselines, including the the relatively simple BERT-CRF. The findings from the main results indicate that the knowledge contained in LLMs can significantly improve sequence labeling performance. However, they lack the necessary expertise and alignment capabilities for handling domain-specific datasets. Due to biases in domain data distribution, LLMs struggle to identify target entities in particular fields as efficiently as conventional sequence labeling techniques.

In addition, we perform LoRA-based SFT on both source and synthetic data, with the results presented in Table~\ref{t10}. For GLM4-9B-Chat and Qwen2.5-14B-Instruct, fine-tuning on synthetic data does not consistently outperform the results reported in Table~\ref{t6}. This suggests that performance is influenced more by training dynamics and stochastic variations during generation and fine-tuning than by the presence of additional synthetic data. Furthermore, we observe a notable performance decline in ChatGLM3-6B and Llama3-8B-Instruct. Prediction analysis reveals that these models tend to overpredict non-target entities, likely because the frequent co-occurrence of entities in both synthetic and original samples makes them overly sensitive to spurious patterns.

Considering the trade-offs between cost and performance, the data augmentation approach leveraging LLMs proposed in this study offers a more practical and efficient solution. This method bridges the gap between LLMs' general knowledge and the specialized requirements of domain-specific sequence labeling tasks.

\section{Results on Full Datasets}\label{a_full}

\begin{table*}[h!]
\resizebox{\textwidth}{!}{%
\begin{tabular}{@{}lcccccccccc@{}}
\toprule
\textbf{Model}      & \textbf{Weibo} & \textbf{Youku} & \textbf{Taobao} & \textbf{Resume} & \textbf{CMeEE} & \textbf{PKU}   & \textbf{MSR}   & \textbf{UD}    & \textbf{CoNLL'03} & \textbf{MIT-Movie} \\ \midrule
W$^2$NER               & 72.59          & 83.62          & 88.27           & 96.88           & 72.97          & 95.51          & 97.72          & 95.00          & 91.71             & 74.62               \\
CNN Nested NER      & 72.31          & 83.79          & 88.86           & 96.67           & 73.83          & 93.75          & 97.69          & 94.88          & 91.16             & 74.86               \\
DiFiNet             & 73.33          & 83.69          & 88.19           & 96.59           & 72.29          & 94.86          & 97.28          & 94.85          & 90.72             & 74.49               \\ \midrule
KnowFREE            & 73.87          & 84.52          & 88.97           & 96.82           & 73.92          & 96.59          & 97.72          & 95.93          & \textbf{92.28}    & 75.32               \\
KnowFREE-F (Deepseek-V3) & \textbf{74.15} & \textbf{84.57} & \textbf{89.12}  & \textbf{96.93}  & \textbf{73.95} & \textbf{96.67} & \textbf{97.76} & \textbf{96.02} & 92.27             & \textbf{75.38}      \\ \bottomrule
\end{tabular}%
}
\caption{Results of sequence labeling tasks on full datasets. The \textbf{bold} values indicate the best performance.}
\label{t9}
\end{table*}

To evaluate the effectiveness of Label Extension Annotation at the full data scale, we conducted additional experiments comparing our method with W$^2$NER, CNN Nested NER, and DiFiNet. The results are shown in Table~\ref{t9}. While performance improvements become less pronounced on certain datasets (e.g., Weibo, Youku, and Resume) compared to the 1000-sample setting, KnowFREE and KnowFREE-F (Deepseek-V3) still outperform all baseline methods on the full datasets, demonstrating their robustness even under high-resource conditions. Although the impact of label extension annotation diminishes as the dataset size increases, it consistently offers improvements over the vanilla KnowFREE, confirming its continued utility. As for enriched explanation synthesis, one of our main motivations was to investigate its performance boundary in low-resource settings. Our analysis shows that its benefits significantly decrease as the sample size grows, with little gain beyond the 500-sample mark. Thus, we can reasonably conclude that enriched explanation synthesis provides limited added value in full-data scenarios.

\section{Cross-Lingual Adaptability}

Our method also extends to other languages, particularly character-dense languages in low-resource settings. To assess its cross-linguistic performance, we conducted additional experiments on Japanese and Korean sequence labeling datasets, as summarized in Table~\ref{t11}.

\begin{itemize}
    \item \textbf{Stockmark-NER (Japanese)}\footnote{https://github.com/stockmarkteam/ner-wikipedia-dataset/}: A Wikipedia-style dataset containing eight entity types, including person, organization, and location names.
    \item \textbf{Naver Changwon NER (Korean)}\footnote{https://github.com/naver/nlp-challenge/tree/master/missions/ner}: A social media dataset containing fourteen entity types, including person names, dates, organizations, and locations.
\end{itemize}

In these experiments, we used GLM4-9B-Chat for data augmentation. The results demonstrate that our method consistently outperforms baseline approaches on both datasets, confirming its effectiveness and adaptability in character-dense, low-resource, multilingual scenarios.

\begin{table}[h]
\caption{Results of sequence labeling tasks on cross-lingual datasets. \textbf{Bold} values indicate the best performance.}
\label{t11}
\resizebox{\columnwidth}{!}{%
\begin{tabular}{l|ccc|ccc}
\toprule
\multirow{2}{*}{\textbf{Model}} & \multicolumn{3}{c|}{\textbf{Stockmark-NER}} & \multicolumn{3}{c}{\textbf{Naver Changwon NER}} \\
                                & 250            & 500            & 1000           & 250              & 500             & 1000            \\ \midrule
W$^2$NER                           & 68.27          & 74.84          & 78.04          & 66.53            & 69.17           & 73.19           \\
CNN Nested NER                  & 69.18          & 75.79          & 79.18          & 68.24            & 70.83           & 74.55           \\
DiFiNet                         & 65.83          & 71.76          & 76.68          & 68.67            & 70.87           & 72.62           \\ \midrule
KnowFREE-F (GLM4-9B-Chat)       & 69.26          & 76.17          & \textbf{79.97} & 68.70            & 70.92           & 75.66           \\
KnowFREE-FS (GLM4-9B-Chat)      & \textbf{73.82} & \textbf{78.12} & 79.87          & \textbf{69.98}   & \textbf{71.23}  & \textbf{75.69}  \\ \bottomrule
\end{tabular}%
}
\end{table}

\section{Impact of the Number of Heads in Local Attention}\label{a_local}

\begin{figure*}[h]
  \includegraphics[width=\textwidth]{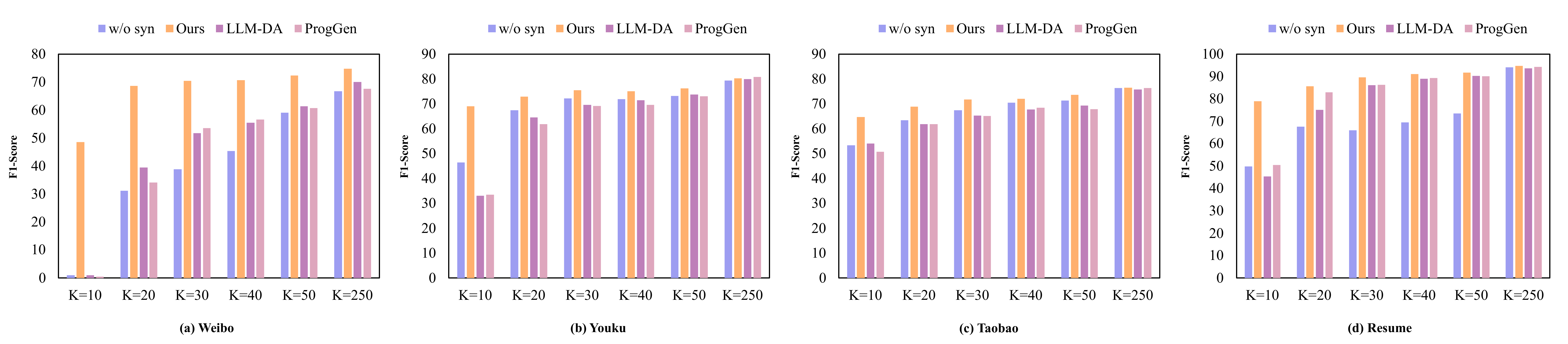}
  \caption{Performance comparison between different data synthesis strategies under k-shot sampling.}
  \label{f8}
\end{figure*}

\begin{figure}[h!]
  \includegraphics[width=0.8\columnwidth]{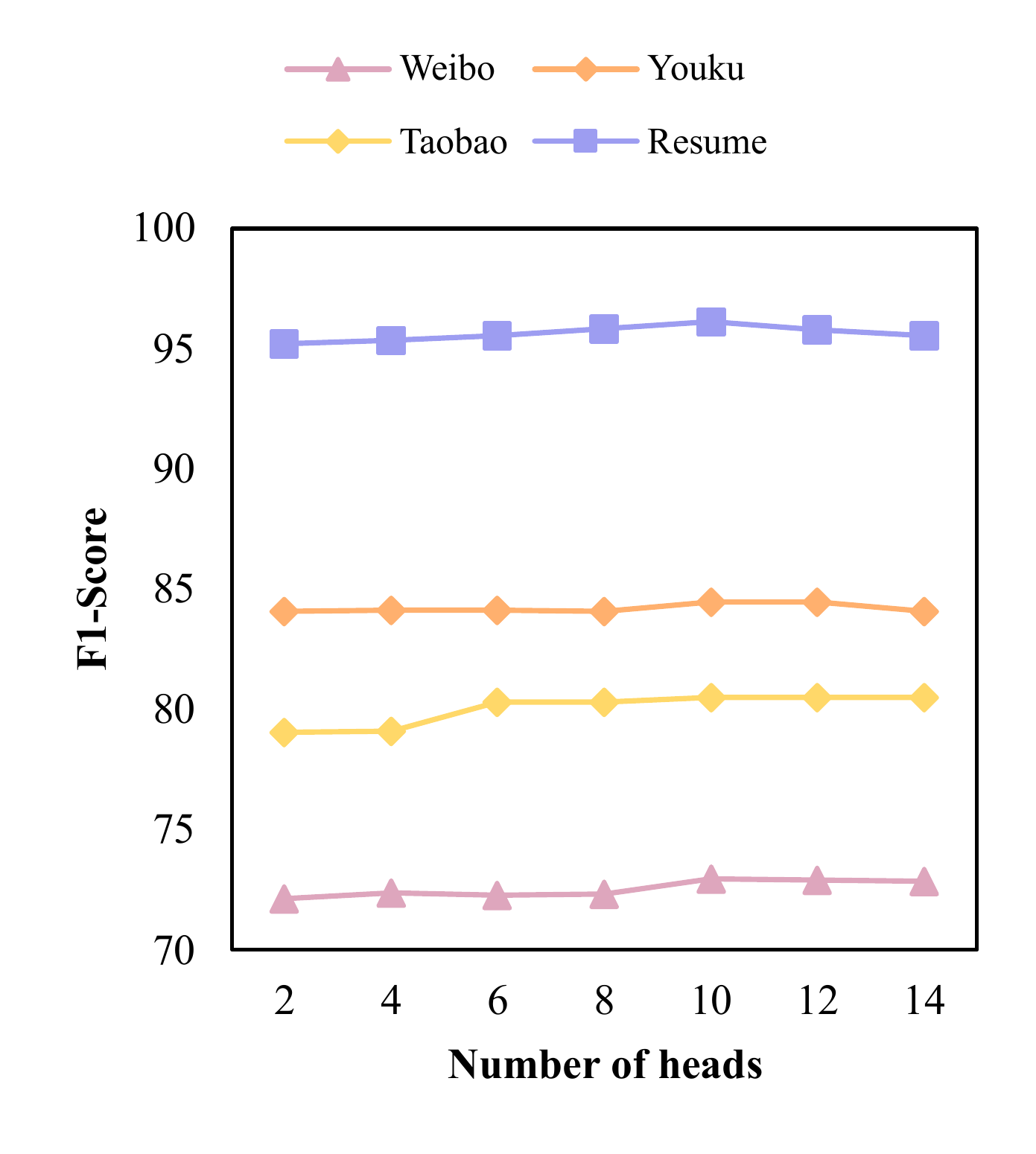}
  \caption{Performance variation with different numbers of heads in the local attention module.}
  \label{f7}
\end{figure}

To analyze the impact of different numbers of attention heads on model performance, we conducted experiments on four flat NER datasets: Weibo, Youku, Taobao, and Resume, using a sampling size of 1000. The model was trained on the sampled data with extension labels extracted by GLM4-9B-Chat, and the results are presented in Figure~\ref{f7}.

The results suggest that the number of attention heads has a relatively moderate influence on model performance. Notably, increasing the number of heads from 8 to 10 yields the most substantial improvement. Moreover, how feature vectors are distributed across heads proves to be a critical factor. To maintain compatibility as the number of heads increases, we adjusted the feature size to ensure it remains divisible by the number of heads. However, this adjustment did not result in further performance gains and significantly increased computational overhead. Therefore, we adopt ten attention heads in this study as a trade-off between performance and efficiency.

\section{Comparison of different data synthesis strategies}\label{a_da}

To evaluate the effectiveness of the enriched explanation synthesis strategy, we reproduced and compared two LLM-based data synthesis methods designed for sequence labeling tasks: LLM-DA and ProgGen. These methods were used to synthesize data from k-shot samples of the original datasets. For consistency, all synthesized samples were annotated using the KnowFREE model trained on the original data without synthesized samples. We conducted experiments on the Weibo, Youku, Taobao, and Resume datasets, all results are presented in Figure~\ref{f8}. The results show that the performance gains from all data synthesis strategies decrease as the number of samples increases. Notably, in scenarios with $\text{k} \leq 30$ on the Youku and Taobao datasets, both LLM-DA and ProgGen lead to performance degradation compared to models trained without synthesized data. This suggests that the synthesized samples generated by these methods may contain inherent semantic distribution biases, which diminish their effectiveness in enhancing performance in certain low-resource domains.

In contrast, our method consistently delivers significantly better performance improvements for $\text{k} \leq 50$, with particularly notable gains on the Weibo dataset. These results demonstrate that the samples synthesized by our approach are more closely aligned with the target domain's distribution and exhibit superior robustness.

\section{Visualization of the Logits with Extension Labels}\label{a_vis}

\begin{figure*}[h]
  \includegraphics[width=\textwidth]{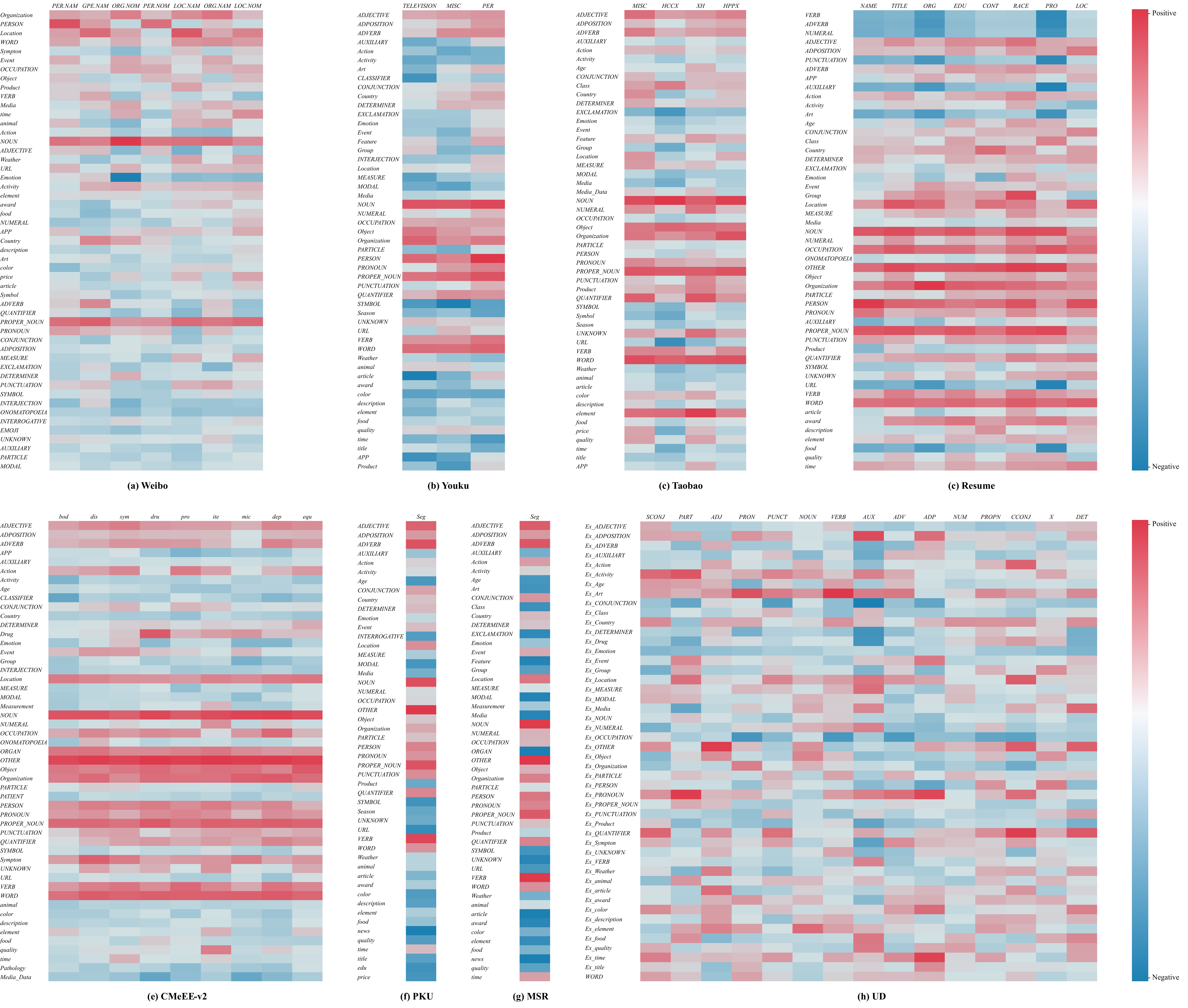}
  \caption{Heatmap visualization of logits scores between target labels and extension labels on the test sets.}
  \label{f6}
\end{figure*}

To further investigate the interaction between the introduced extension labels and target labels in the model, we visualized the logit scores of extension labels corresponding to each predicted target label position in the test set. These scores were aggregated by summing and averaging across label categories, and the results are displayed as a heatmap in Figure~\ref{f6}. In each heatmap subplot, the horizontal axis represents the target labels, while the vertical axis corresponds to the extension labels.

The results reveal that certain extension labels exhibit strong correlations with specific target labels. For instance, in the Weibo dataset, both ``PER.NAM'' and ``PER.NOM'' show a notable association with the extension label ``PERSON''. Similarly, in the Resume dataset, ``COUNT'' demonstrates strong correlations with the extension labels ``Country,'' ``Location,'' and ``description''. Incorporating these significant relationships during training allows the model to leverage co-occurrence patterns, enhancing its ability to perform fine-grained semantic understanding and improving target entity prediction.

However, some extension labels were observed to have strong correlations with all target labels. Since the KnowFREE model produces independent probability scores for each target label, the influence of such extension labels on target entity predictions is generally limited when their number is small, as their training weights are relatively low. Conversely, when these extension labels become overly numerous, they can negatively impact model training. In such cases, reducing their weights further can help mitigate these effects and enhance overall model performance.

\section{Statistics of Datasets}
\label{sta}

\begin{table}[h]
\centering
\resizebox{\columnwidth}{!}{%
\begin{tabular}{lcccl}
\toprule
\textbf{Dataset} & \textbf{Dev} & \textbf{Test} & \textbf{Label Types} & \textbf{Domain}    \\ \midrule
Weibo            & 0.27k        & 0.27k         & 8                    & Social Media     \\
Youku            & 1.00k        & 1.00k         & 3                    & Video Content    \\
Taobao           & 1.00k        & 1.00k         & 4                    & E-commerce         \\
Resume           & 0.46k        & 0.48k         & 8                    & Human Resources  \\
CMeEE-v2          & 4.98k        & 4.98k         & 9                    & Medical            \\
PKU              & 1.00k        & 2.04k         & 1                    & News               \\
MSR              & 1.00k        & 3.99k         & 1                    & News               \\
UD               & 0.50k        & 0.50k         & 16                   & News, Literature \\
CoNLL'03          & 3.47k        & 3.68k         & 4                    & News            \\
MIT-Movie          & 1.00k        & 1.95k         & 12                    & Entertainment            \\ \bottomrule
\end{tabular}%
}
\caption{Statistics of development sets, test sets, label types and domains of all datasets.}\label{t4}
\end{table}

The detailed statistics of the datasets are shown in Table~\ref{t4}. These datasets span various domains, including Social Media, E-commerce, and Medical, enabling a comprehensive evaluation of the model's performance across different fields.

\section{More Experiment Settings}\label{a_set}

In this section, we describe the additional experimental parameter settings for our method. In the pipeline of label extension annotation, the pre-trained embedding model of $\mathcal{M}$ is set as ``text2vec'' \citep{text2vec}, the Top-$p$ value is set as five, and the threshold $\epsilon$ is set as 1.5. In the training stage, the hidden size $D$, $D^\prime$, and $\tilde{D}$ are set as 768, 200, and 200, respectively. The activation function of $\sigma$ and $\sigma^{\ast}$ are defined as Leaky ReLU and GeLU, respectively. In the local multi-head attention module, the window size $\omega$ is set as three and the number of attention heads $\mathcal{K}$ is set as ten. In the sequence labeling model, we distinguish between the learning rate for the PLM and other modules, setting them to 2e-5 and 1e-3, respectively. For the weight $\alpha$ of extension labels, we introduce a dynamic weight calculation mechanism to handle the influence of frequently occurring extension labels (e.g., POS tags). These frequent labels can affect the gradient calculation, leading to reduced attention to target labels. To address this, we calculate the count $C_i$ of each extension label and the average count $\hat{C}$ of target labels, and then compute the weight coefficient $\alpha_i$ as follows: 
\begin{equation}
  \alpha_i = 0.5 \times (\hat{C}/C_i).
\end{equation}

\begin{table}[h]
\resizebox{\columnwidth}{!}{%
\begin{tabular}{lccl}
\toprule
\textbf{Dataset} & \textbf{Weight Decay} & \textbf{$\beta$} & \textbf{Type}  \\ \midrule
Weibo            & 1e-3                  & 1.0                           & Flat   NER     \\
Youku            & 1e-2                  & 0.4                           & Flat   NER     \\
Taobao           & 1e-3                  & 0.4                           & Flat   NER     \\
Resume           & 1e-3                  & 0.4                           & Flat   NER     \\
CMeEEv2          & 1e-3                  & 0.4                           & Nested   NER   \\
PKU              & 1e-2                  & 1.0                           & Word   Segment \\
MSR              & 1e-2                  & 1.0                           & Word   Segment \\
UD               & 1e-2                  & 1.0                           & POS   Tagging  \\
CoNLL'03               & 1e-3                  & 0.4                           & Flat   NER  \\
MIT-Movie               & 1e-2                  & 0.4                           & Flat   NER  \\\bottomrule
\end{tabular}%
}
\caption{Settings of $\beta$ and weight decay across different datasets.}\label{t7}
\end{table}

The training weight for synthesized samples ($\beta$) and the weight decay parameter are provided in Table~\ref{t7}. As shown, for most NER datasets with more complex entity semantics, we use a smaller weight decay parameter to improve model fitting during training. In contrast, for POS tagging and tokenization datasets, we apply a larger weight decay to prevent overfitting. Additionally, for NER datasets, where entity labels are more prone to noise from synthesized samples, we set $\beta = 0.4$. On the other hand, for datasets with strong baseline performance, increasing $\beta$ to 1.0 helps the model better utilize synthesized samples during training. Our implementation is built on the Huggingface Transformers \citep{wolf-etal-2020-transformers}, and all experiments are conducted using two NVIDIA A6000 GPUs for both training and inference.

\section{Prompts}

\begin{figure*}[t!] 
  \centering
  \begin{minipage}{0.48\textwidth} 
      \centering
      \includegraphics[width=\linewidth]{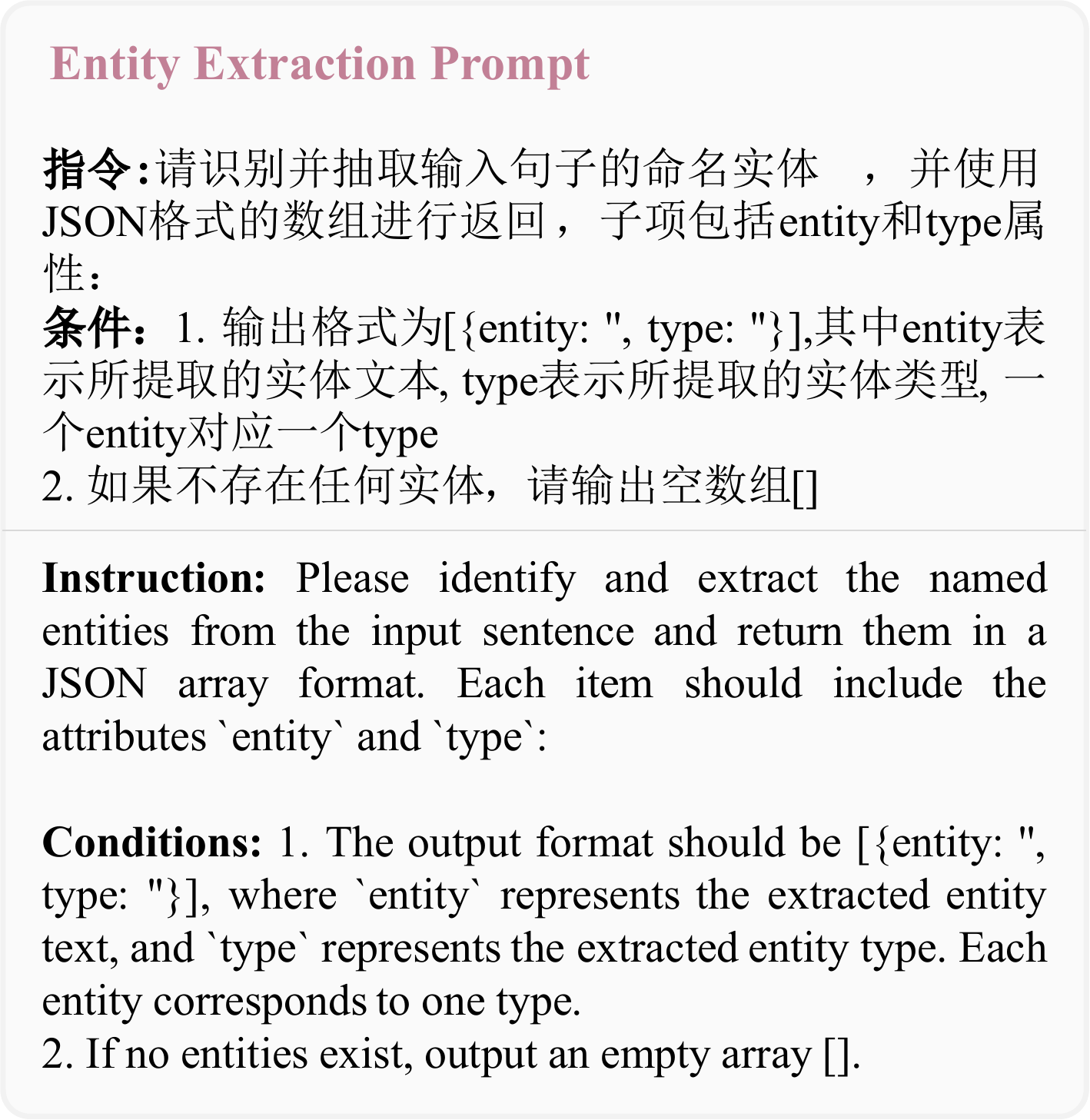}
      \caption{The entity extraction prompt in label extension annotation.}
      \label{fp3}
  \end{minipage}
  \hfill 
  \begin{minipage}{0.48\textwidth} 
      \centering
      \includegraphics[width=\linewidth]{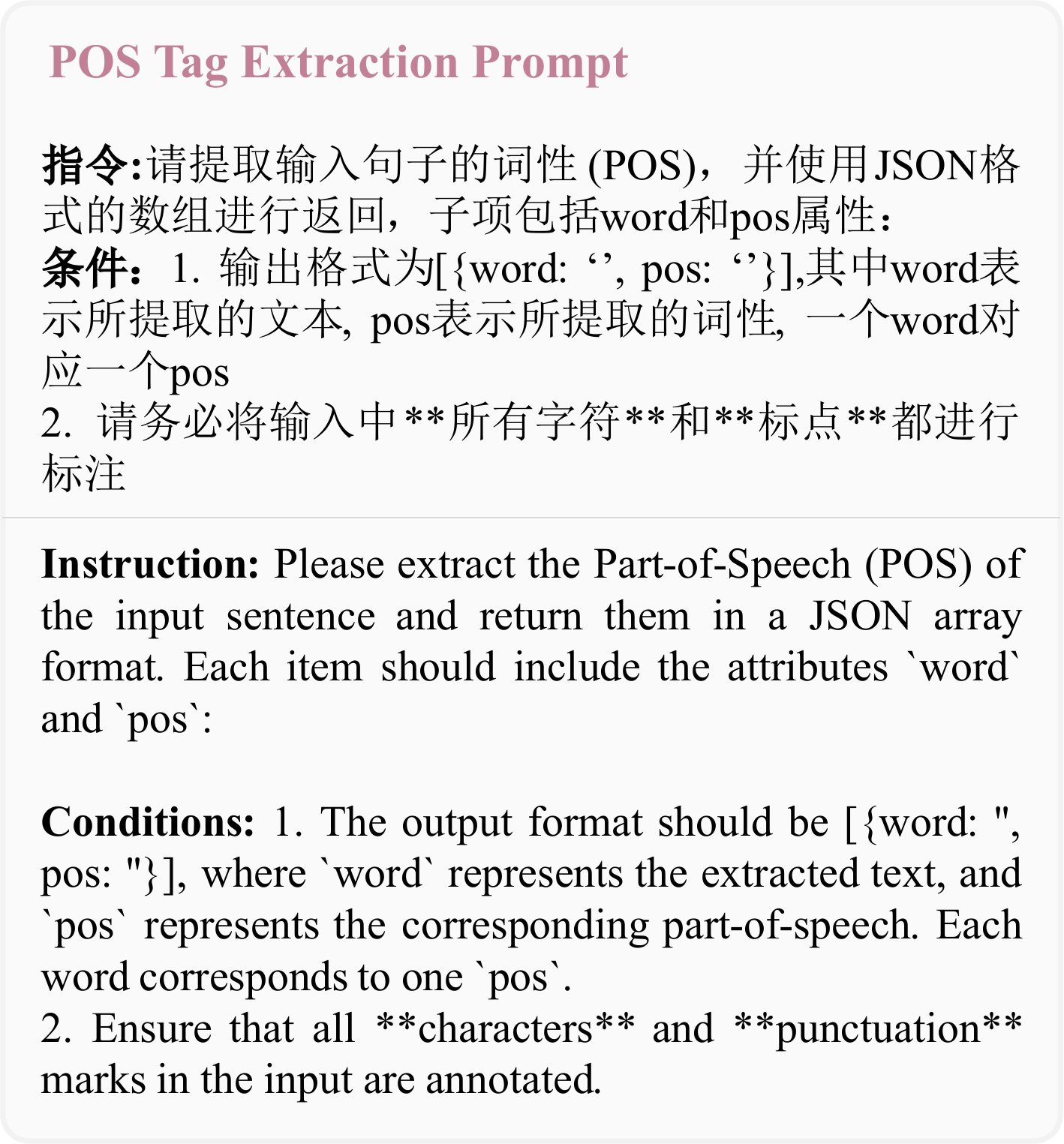}
      \caption{The POS tag extraction prompt in label extension annotation.}
      \label{fp4}
  \end{minipage}
\end{figure*}

\begin{figure*}[t!] 
  \centering
  \begin{minipage}{0.48\textwidth} 
      \centering
      \includegraphics[width=\linewidth]{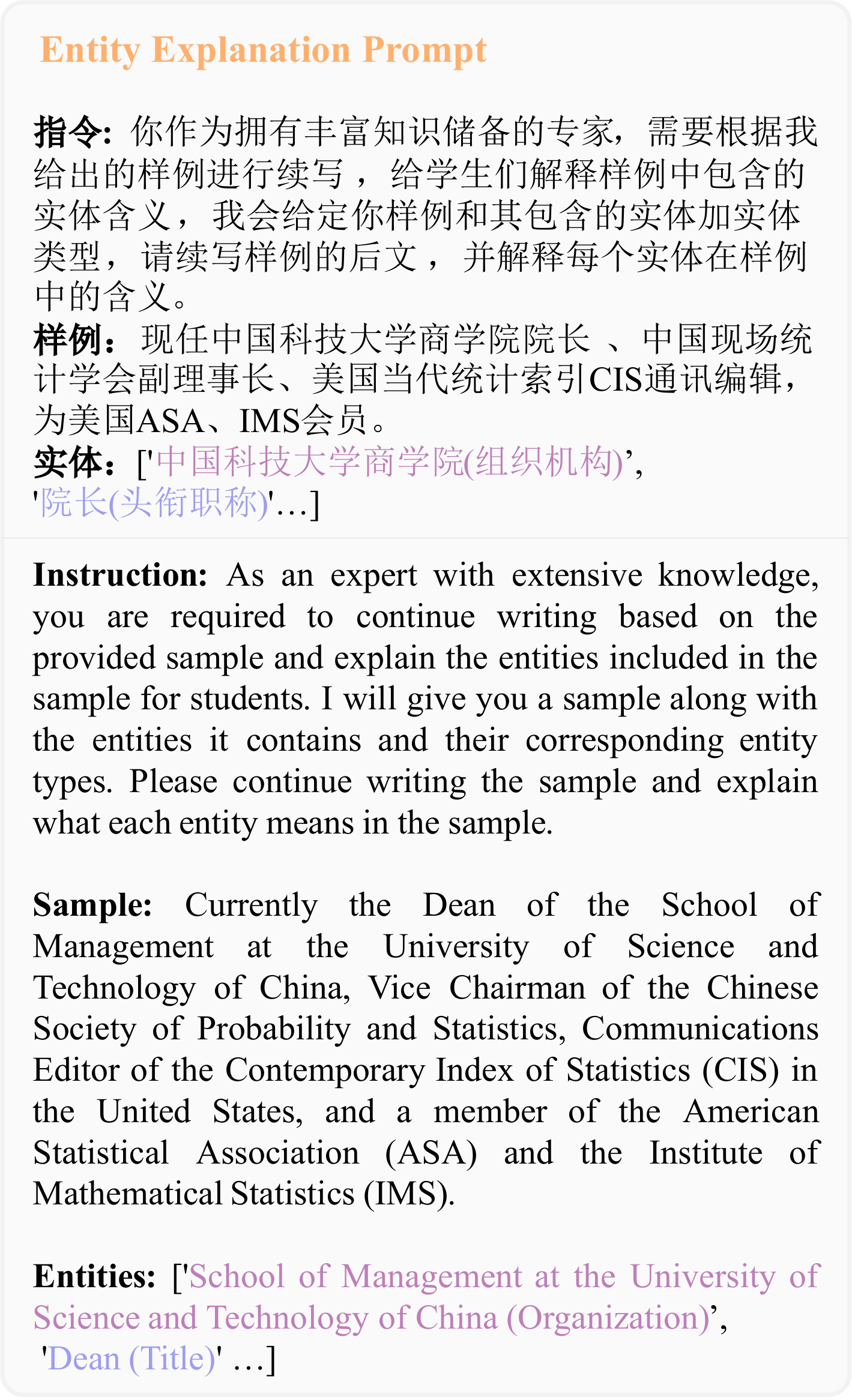}
      \caption{The entity explanation prompt in enriched explanation synthesis.}
      \label{fp1}
  \end{minipage}
  \hfill 
  \begin{minipage}{0.48\textwidth} 
      \centering
      \includegraphics[width=\linewidth]{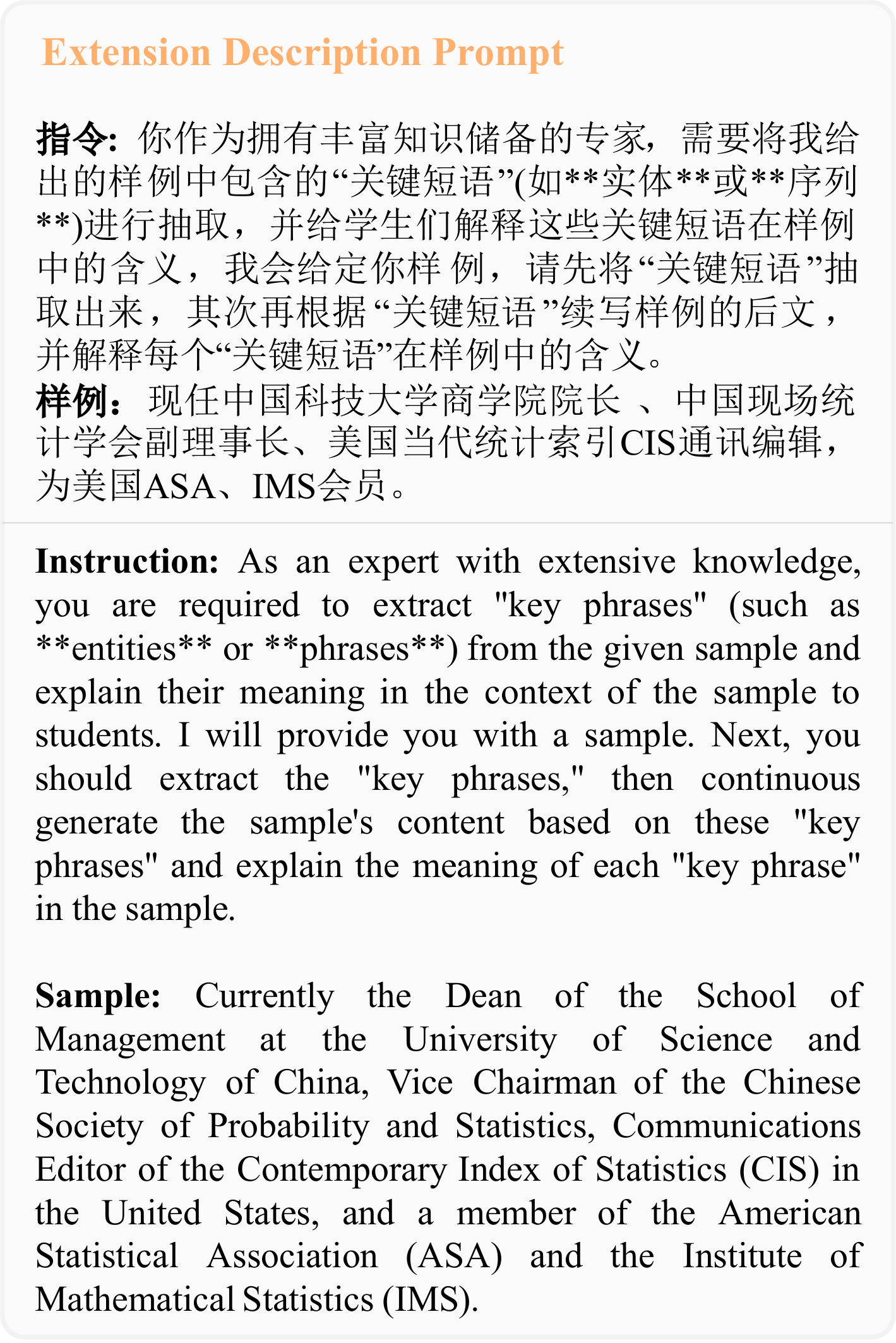}
      \caption{The extension description prompt in enriched explanation synthesis.}
      \label{fp2}
  \end{minipage}
\end{figure*}

In this section, we present detailed examples of our workflow prompts for label extension annotation in Figure~\ref{fp3}, \ref{fp4} and enriched explanation synthesis in Figure~\ref{fp1}, \ref{fp2}. Since the target dataset is entirely in Chinese, all original prompts are written in Chinese. The English portions in the prompt examples are translations of the original prompts.

\end{document}